\ifdefined\pdfminorversion\pdfminorversion=7\fi
\documentclass{article}

\PassOptionsToPackage{numbers,compress}{natbib}

\usepackage[preprint]{neurips_2026}

\usepackage[utf8]{inputenc}
\usepackage[T1]{fontenc}
\usepackage{graphicx}
\usepackage{tikz}
\usetikzlibrary{arrows.meta,positioning}
\usepackage{booktabs}
\usepackage{tabularx}
\usepackage{amsmath,amssymb}
\usepackage{algorithm}
\usepackage{algpseudocode}
\usepackage{nicefrac}
\usepackage{microtype}
\usepackage{xcolor}
\usepackage{url}
\usepackage{hyperref}
\hypersetup{hidelinks,pdftitle={CondenseVLA: Learnable History Condensation for Efficient Multi-Frame VLA},
  pdfauthor={Feiyang Hong, Yujie Wei, Bo Zhao, Xiu Su, Hongxun Yao, Shuo Yang}}
\usepackage[capitalize,nameinlink]{cleveref}

\makeatletter
\renewcommand{\theHALG@line}{\thealgorithm.\arabic{ALG@line}}
\makeatother

\newcommand{\MethodName}{CondenseVLA}
\newcommand{\ModuleName}{Learnable History Condensation}
\newcommand{\ModuleShort}{LHC}

\title{CondenseVLA: Learnable History Condensation for Efficient Multi-Frame VLA}

\author{%
  Feiyang Hong$^{1}$ \hspace{0.6em} Yujie Wei$^{2}$ \hspace{0.6em} Bo Zhao$^{3}$ \hspace{0.6em} Xiu Su$^{4}$ \hspace{0.6em} Hongxun Yao$^{2}$ \hspace{0.6em} Shuo Yang$^{1}$\thanks{Corresponding author.\newline Emails: \texttt{feiyanghong7040@gmail.com}, \texttt{shuoyang@hit.edu.cn}.} \\[0.4em]
  {\normalfont $^{1}$ Harbin Institute of Technology, Shenzhen} \\
  {\normalfont $^{2}$ Harbin Institute of Technology} \\
  {\normalfont $^{3}$ Shanghai Jiao Tong University \quad $^{4}$ Central South University}
}

\begin{document}

\maketitle
\begin{abstract}
Robot manipulation often depends on short-term cues---such as brief contacts, temporary occlusions, and intermediate object states---that are ambiguous in a single frame and can only be resolved by integrating observations over time. However, injecting historical information solely by enlarging the history window is computationally costly and frequently yields limited gains due to substantial redundancy and task-agnostic, inflexible sampling of observations.
We propose \MethodName, a multi-frame VLA policy with a learnable decoder-side history condensation module for using short-term visual context in action decoding. Given a sequence of cached per-frame features from the VLM backbone, \MethodName\ distills observation history into a fixed set of slot tokens using learnable queries and multi-head cross-attention. Simple auxiliary regularization penalizes overlap among slot attention maps and decorrelates slot embeddings, encouraging complementary attention and reducing slot redundancy.
Across the evaluated tasks, \MethodName\ improves Simpler-WidowX success from 58.7\% to 69.8\% (+11.1 percentage points), achieves 97.2\% on LIBERO, and reaches 52.8\% on real-world tasks. It achieves 21.6\% success on memory-dependent RMBench tasks. These results support selective condensation of bounded recent visual evidence for action decoding, while harder multi-observation tasks expose the limits of a finite history window.
Project page: \url{https://github.com/FeiyangHong/CondenseVLA}
\end{abstract}

\section{Introduction}
\label{sec:intro}

\begin{figure}[t]
  \centering
  \includegraphics[width=0.98\linewidth]{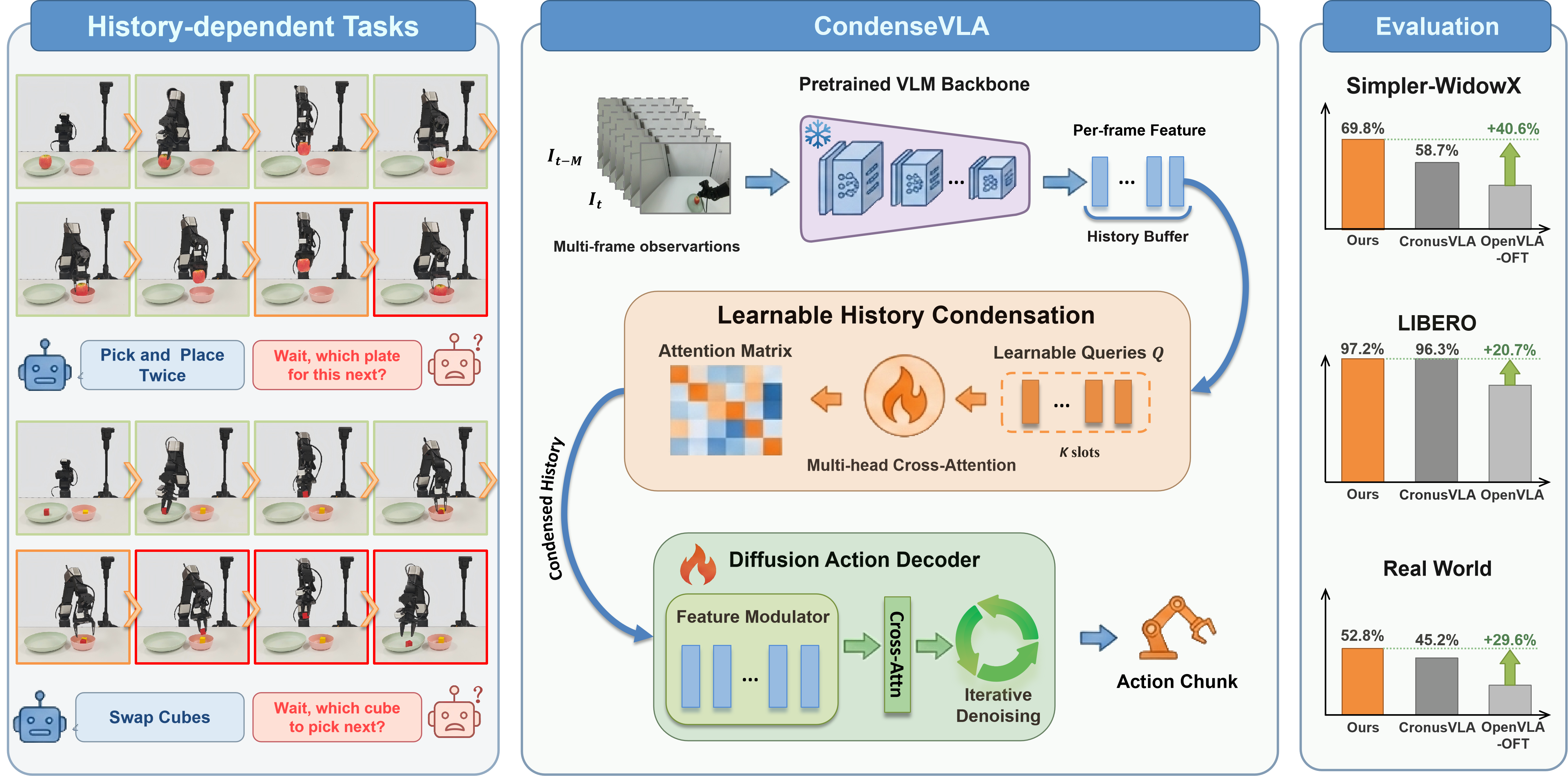}
  \caption{We propose \MethodName, a decoder-side \ModuleName\ (\ModuleShort) model that distills informative multi-frame context into compact slot tokens. The left panel highlights history-dependent failures of single-step VLA execution: in Pick and Place Twice, the policy forgets which plate to place next; in Swap Cubes, it forgets which cube should be picked next. Across multiple benchmarks, including Simpler, LIBERO, and real-world robot experiments, \MethodName\ consistently outperforms the CronusVLA baseline.}
  \label{fig:teaser_overview}
\end{figure}

Robot manipulation tasks, especially long-horizon and partially observable ones, require temporal reasoning over contact events, state transitions, and other transient visual cues. For example, in Pick and Place Twice, the current frame may only show the gripper holding an object, but not whether the next correct action is to place it on the other plate or return it to the original one. Robust execution therefore depends on integrating observations over time rather than reacting to a single frame. Vision-language-action (VLA) policies have recently made significant progress in multi-task generalization \cite{brohan2023rt1,Kim2024}. Single-frame policies such as OpenVLA learn diverse skills from large-scale robot trajectory datasets but lack explicit observation history \cite{Kim2024}. RT-2 integrates a web-scale vision-language model to enable zero-shot reasoning, advancing general-purpose robotic control \cite{brohan2023rt2}. However, single-frame conditioning cannot directly access earlier visual observations needed to resolve task progress \cite{Li2025}. A naive solution is to increase the history window by stacking more frames; however, this quickly becomes computationally expensive and increases latency---as reported in HAMLET, appending only four additional past observation frames can slow the forward pass by about 35\% \cite{Koo2025HAMLET}---and often brings little or even negative gain because nearby frames are highly repetitive and fixed-interval sampling can miss short but important task events. When irrelevant or repetitive observations dominate the context, longer histories can inject noise and degrade policy performance. This motivates a more intelligent way to encode temporal context, filtering redundancy while preserving important temporal cues.

Recent work has begun exploring efficient temporal context for robotic policies~\cite{qian2026eida}. Representative examples such as ContextVLA and CronusVLA show that history-aware modeling can substantially improve long-horizon robustness while controlling computational overhead \cite{Jang2025,Li2025}. These advances highlight that temporal context is critical for robust VLA manipulation, but they also expose a central challenge: how can short-term temporal context be incorporated into VLA policies in a computationally efficient and robust manner~\cite{wang2026flashnav}?

As illustrated in \cref{fig:teaser_overview}, history-dependent tasks such as Pick and Place Twice and Swap Cubes require recalling previously completed sub-steps, which is difficult for single-step conditioning. We therefore propose \ModuleName\ (\ModuleShort), which encodes a length-$M$ observation history into $K$ context tokens ($K \ll M$). The \ModuleShort\ module is inserted between the pretrained vision--language backbone and the diffusion action decoder \cite{Chi2024DiffusionPolicy,Peebles2023DiT}. At each time step, the VLM encodes only the current observation; the most recent $M$ historical features are reused from a FIFO cache and condensed into $K$ slot embeddings. The current feature is supplied separately to the decoder.
We instantiate the \ModuleShort\ module with cross-attention: $K$ learnable slot queries attend to the $M$ historical features as keys and values to extract instruction-relevant information. The learned slots provide compact conditioning, with attention visualizations suggesting complementary coverage of the recent history; slot indices do not prescribe particular events or temporal positions. Conceptually, our design follows the spirit of slot-attention methods that condense variable-length sets into latent slots, and adapts it to distill multi-frame observations into compact context tokens.

Our main contributions are summarized as follows:
\begin{itemize}
\item Method. We propose a decoder-side Learnable History Condensation mechanism for multi-frame VLA policies, which distills cached recent VLM features into a fixed number of action-conditioning slots; auxiliary regularization encourages complementary attention and reduces slot redundancy.
\item Empirical results. \MethodName\ improves Simpler-WidowX success from 58.7\% to 69.8\%, achieves 97.2\% on LIBERO, reaches 52.8\% on real-world tasks, and achieves 21.6\% on RMBench.
\item Findings. We investigate why history condensation matters. Our results show that long histories contain substantial redundancy, and that learning task-relevant history condensation is more reliable than naively expanding history windows or rigid sampling of observations.
\end{itemize}

\section{Related Works}

\subsection{Vision-Language-Action Models}
Recent research on VLA models has produced large-scale foundation policies trained on a combination of internet-scale vision-language corpora and robot data~\cite{Li2024c,zhou2026inject}. OpenVLA is built on a LLaMA2-based language backbone with pretrained visual encoders \cite{Kim2024}. It exhibits strong multi-task manipulation generalization and compares favorably with prior closed-source models such as RT-2 (55B) \cite{brohan2023rt2}. GR00T N1 adopts a dual-system design that couples a vision-language module for high-level reasoning with a diffusion Transformer module for low-level motion control, and is jointly trained on diverse real-robot data, human videos, and synthetic data \cite{Bjorck2025}. Beyond general-purpose foundation models, specialized VLA variants have also emerged. Spatially enriched VLA models, such as SpatialVLA and DepthVLA, incorporate 3D-related cues to improve geometric grounding and robustness across camera viewpoints \cite{Qu2025,zhang2025spatial,yuan2025depthvla,li2025qdepth}. Related work on unsupervised 3D scene understanding learns semantic structure through superpoint grouping and growing semantic primitives~\cite{zhang2025logosp,zhang2026growsp++}. VLA models with force/torque or tactile signals, such as TA-VLA and Tactile-VLA, further improve contact-aware decision making in manipulation \cite{huang2025tactile,zhang2025elucidating}. TouchWorld combines tactile prediction with reactive control, while TouchAnything estimates bimanual tactile signals from egocentric video~\cite{zhou2026touchworld,zhou2026touchanything}. Experience-driven VLA models, such as HAVE and DEJAVU, leverage historical interaction traces to improve temporal consistency and recovery from failures \cite{lee2025imagine,li2025learn,li2025reflection,wu2025dejavu}. DEWO adapts world-action models using visual supervision from successful and failed deployment continuations~\cite{zhan2026directexperienceworldmodeloptimization}. Meanwhile, EgoVLA scales training with large amounts of egocentric human video, learning to predict human hand actions from video and then retargeting them to robots \cite{Yang2025}. ConLA learns latent actions from human videos through contrastive disentanglement, and WING transfers interaction-centric latent actions using spectral guidance~\cite{dai2026conla,liu2026worldactionlearninginteractioncentric}. Recent studies also investigate target-aware and cognition-aware attention optimization in VLA, as exemplified by CogVLA, and enhance VLA with explicit perceptual-cognitive memory mechanisms, as in MemoryVLA; $\pi_{0.5}$ further strengthens open-world generalization \cite{li2025cogvla,li2026SemanticVLA,shi2025memoryvla,Black2025pi05}. Collectively, these efforts reflect a broader trend toward robotic foundation models, leveraging vision-language priors and high-quality robot data to enable broad manipulation capabilities \cite{Driess2023PaLME,guo2026utilgen}.

\subsection{Multi-frame Modeling and Efficient Inference}
Temporal VLA methods differ in where they summarize history and what structure they retain. TraceVLA overlays past motion cues on the current observation~\cite{Zheng2024}, while CronusVLA uses two-stage training to extend single-frame OpenVLA with dense multi-frame conditioning~\cite{Li2025}. EgoVLA instead exploits temporal demonstrations from human video for transfer to robot control~\cite{Yang2025}.

ContextVLA average-pools historical hidden states into one context token at an intermediate VLM layer; later backbone layers process this token with the current observation, while causal masking enables reuse of cached history~\cite{Jang2025}. It is trained with the underlying policy's action objective. \MethodName\ instead retains frozen vision/language weights and maps cached frame features to $K$ learned summaries at the diffusion-decoder interface. Multiple slots provide an adjustable conditioning budget, at the cost of added parameters and choosing $K$; their number alone does not establish an advantage over a single token. HAMLET appends several moment tokens to each VLM input, initializes them with time-contrastive learning, and uses a shallow causal Transformer to aggregate recent moment tokens before action prediction~\cite{Koo2025HAMLET}. Our LHC is jointly optimized with the action decoder without a separate moment-token initialization stage. Both use recent temporal summaries, but differ in backbone interaction and temporal aggregation.

SlotVLA learns object slots and relation tokens from dense visual patches. Its two-stage training uses box, mask, and task-relevance supervision before relation/action learning, and carries object slots between frames to preserve identity~\cite{Hanyu2025}. Its slots therefore encode supervised object structure, whereas our slots aggregate cached instruction-conditioned frame features without object labels or recurrent object-slot state. HistRISE segments task-relevant objects, tracks their points in 3D, and compresses variable-length tracks into one history token per object for policy conditioning~\cite{Chen2025}. It can use full or pre-loaded histories, but relies on segmentation and point tracking. Our fixed FIFO avoids these additional perception components while relinquishing an explicit representation of events outside the recent window. These are differences in representation, supervision, and memory scope, not evidence of superiority under matched conditions.

Token merging~\cite{Choi2024}, computation reuse~\cite{Xu2025VLACache,Yang2025EfficientVLA}, and action tokenization~\cite{Pertsch2025} address complementary efficiency concerns. TaLo identifies task-interfering VLM layers and bypasses a selected layer at inference~\cite{liu2026all}. At a different level, MemER retrieves historical keyframes through a hierarchical policy~\cite{Sridhar2026MemER}, and MEM combines short-horizon video memory with text-based task memory~\cite{torne2026mem}. \MethodName\ focuses on policy-oriented condensation of recent temporal cues at the decoder interface. We use \ModuleName\ to distill cached per-step VLM features into several slot tokens, directly controlling the trade-off between temporal horizon and slot capacity.

\section{Preliminaries}

\subsection{Multi-frame in VLA Policies}
Multi-frame manipulation can be viewed as a partially observable decision process in which actions depend on latent task state inferred from an observation history, rather than from a single frame. A common design principle is to decouple per-frame semantic encoding from temporal aggregation: a vision-language backbone extracts frame-level features, and a lightweight temporal module integrates them for control. This factorization preserves strong visual-language priors while keeping temporal reasoning computationally tractable. Existing systems instantiate this principle in different ways. For example, CronusVLA adopts a two-stage pipeline (single-frame pretraining followed by multi-frame post-training with a diffusion decoder) \cite{Li2025}, while ContextVLA amortizes history into reusable context tokens with caching for efficient inference \cite{Jang2025}.

\subsection{Temporal Feature Buffering and Decoding}
Let $o_t$ denote the RGB observation at time $t$ and $x$ the language instruction. A pretrained VLM backbone maps $(o_t,x)$ to a frame-level feature $f_t \in \mathbb{R}^D$. A history window is represented as
$F_t^{(M)}=\{f_{t-M},\ldots,f_{t-1}\}\in \mathbb{R}^{M\times D}$,
which is typically maintained as a FIFO feature buffer so past features can be reused and the VLM is evaluated only once per step. The action decoder (e.g., a diffusion decoder) consumes a fixed-length sequence derived from $\{f_{t-M},\ldots,f_t\}$ and predicts an action chunk $a_{t:t+T_a-1}$. This creates a core multi-frame trade-off: increasing $M$ improves temporal coverage but also increases redundancy and computation. Therefore, multi-frame methods require a history-to-token mapping that selects or condenses informative temporal information into a bounded number of tokens.

\section{Method}
\label{sec:method}
\begin{figure}[t]
  \centering
  \includegraphics[width=0.98\linewidth]{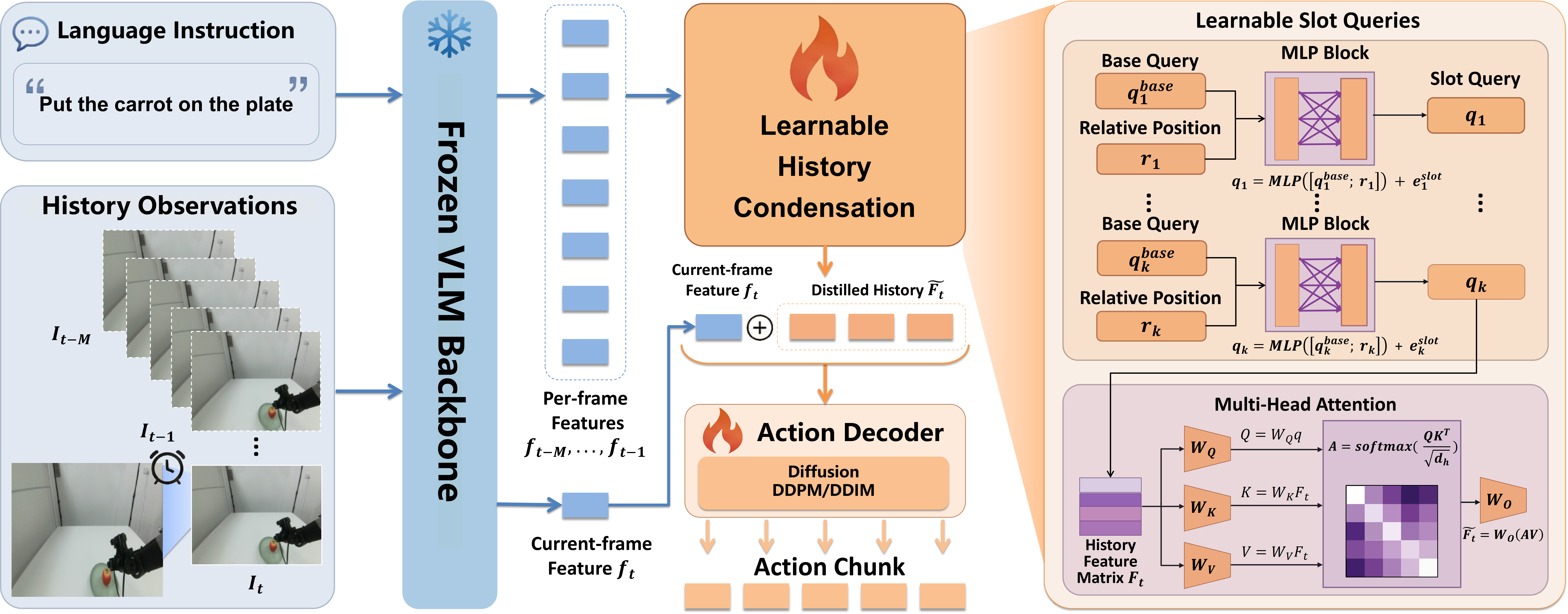}
  \caption{Overview of \MethodName. Frozen vision/language weights and a trainable multimodal projector encode each observation. LHC condenses cached history using slot queries and value-only time embeddings. The decoder expands the current feature separately; historical features use stop-gradient during training.}
  \label{fig:cronus_selector_overview}
\end{figure}

The proposed method condenses per-frame VLM features in a multi-frame VLA pipeline. At step $t$, the VLM encodes the current observation into $f_t \in \mathbb{R}^D$, a FIFO buffer stores $M$ history features $F_t=[f_{t-M},\ldots,f_{t-1}]\in \mathbb{R}^{M\times D}$, and \ModuleShort\ condenses them into $K$ history slots $\tilde{F}_t\in \mathbb{R}^{K\times D}$ ($K \ll M$). The diffusion decoder is conditioned on the fixed-length sequence $Z_t=[\tilde{F}_t;f_t]\in \mathbb{R}^{(K+1)\times D}$. By placing temporal compression at the action-decoder interface, the decoder receives $K$ temporal conditioning tokens independent of $M$, while past frames are encoded once, cached, and selectively summarized before action decoding. An overview is shown in \cref{fig:cronus_selector_overview}.

\subsection{Learnable History Condensation (\ModuleShort)}
\ModuleShort\ maps $F_t$ to $K$ slot tokens via multi-head attention with learnable Q/K/V projections. Although the slot queries are learned parameters, they attend over instruction-conditioned VLM history features, allowing the resulting slot assignments to capture task-relevant temporal evidence under a fixed decoder-side token budget. Each slot $i$ maintains a base query $q_i^{\text{base}} \in \mathbb{R}^D$, a slot embedding $e_i^{\text{slot}} \in \mathbb{R}^D$, and a fixed relative scalar $r_i\in[0,1]$ injected through an MLP:
\begin{equation}
q_i=\mathrm{MLP}([q_i^{\text{base}};r_i])+e_i^{\text{slot}},\quad i=1,\ldots,K.
\end{equation}
The queries $q=[q_1;\ldots;q_K]$ attend to cached history with $H$ heads of width $d_h=D/H$. Learned history-position embeddings enter only the value path, with a learnable scale initialized to zero. For head $h$, let $Q^{(h)}$, $K_{\mathrm{hist}}^{(h)}$, and $V^{(h)}$ denote the projected queries, keys, and values:
\begin{equation}
A^{(h)}=\mathrm{softmax}\left(\frac{Q^{(h)}(K_{\mathrm{hist}}^{(h)})^{\top}}{\sqrt{d_h}}\right)\in \mathbb{R}^{K\times M}.
\end{equation}
\begin{equation}
\tilde{F}_t=\mathrm{MHA}(q,F_t)=\mathrm{ConcatHeads}\!\left(A^{(h)}V^{(h)}\right)_{h=1}^{H}W_O\in \mathbb{R}^{K\times D}.
\end{equation}
Softmax normalizes over history positions, and $W_O$ projects the concatenated heads. Full projections and tensor dimensions are given in the appendix.
No hard temporal assignment is imposed on the output slots. Instead, the relative scalar only breaks permutation symmetry and provides a weak slot identity, allowing slot roles to emerge from end-to-end training.

\subsection{Injecting Condensed History into the Diffusion Decoder}
The decoder input is $Z_t=[\tilde{F}_t;f_t]$, where $\tilde{F}_t$ provides compact temporal evidence and $f_t$ anchors the latest scene state. In cross-attention, the history slots and current-step feature play complementary roles: the slots preserve information about past contacts, object states, and subgoal progress, while the current feature grounds action generation in the latest observation. This fixed-length conditioning avoids passing dense, repetitive history tokens to the diffusion decoder, which is important when increasing the history window would otherwise increase latency and introduce redundant visual evidence. Thus, history affects action prediction through conditioning tokens rather than raw sequence re-encoding.

\subsection{Training Objective}
Fine-tuning optimizes the multimodal projector, LHC, and diffusion decoder while freezing the vision encoders and language model. The total objective is
$L=L_{\mathrm{diff}}+\lambda_{\mathrm{div}}L_{\mathrm{div}}+\lambda_{\mathrm{slot}}L_{\mathrm{slot}}.$

Diffusion training follows standard noise prediction for the noised action chunk $a^{\epsilon}$:
\begin{equation}
L_{\mathrm{diff}}=\mathbb{E}_{\epsilon,\tau}\left[\mathrm{MSE}\!\left(\epsilon,\epsilon_{\theta}(a^{\epsilon},Z_t;\tau)\right)\right].
\end{equation}
Here $\epsilon\sim\mathcal{N}(0,I)$, $\tau$ is the diffusion timestep, and MSE averages squared errors over all entries.

Attention diversity uses the head- and mini-batch-averaged, row-normalized attention $\bar A$:
\begin{equation}
L_{\mathrm{div}}=\mathrm{MSE}(\bar A\bar A^{\top},I).
\end{equation}

Off-diagonal terms penalize overlap; diagonal terms encourage concentrated rows. Slot decorrelation averages squared cosine similarity over selected pairs:
\begin{equation}
L_{\mathrm{slot}}=\operatorname{Mean}_{b,(i,j)\in\mathcal P}\!\left[\cos^2(\tilde f_{b,i},\tilde f_{b,j})\right].
\end{equation}
We use middle-only pairs $\mathcal P=\{(i,j):2\leq i<j\leq K-1\}$. The exact loss reductions, denominator protection, and weighted variants are given in Appendix~\ref{sec:appendix_loss_reductions} and Eq.~\ref{eq:slot_weighted}.

\subsection{Fine-tuning Protocol and Computational Cost}
At inference, a FIFO caches frame features and the VLM is evaluated once per environment step. The LHC attention-mixing operation costs $\mathcal{O}(KMD)$. The $K+1$ input features become $K+6$ decoder-conditioning tokens after current-feature expansion, independent of $M$. LHC has 100.8M parameters, about 1.35\% of the reported 7.47B frozen vision/language backbone and 74.8\% of the 134.7M decoder. Appendix~\ref{sec:appendix_protocol_details} summarizes the architecture and parameter counts.

\section{Experiments}

\subsection{Results in Simulation}

\subsubsection{Implementation Details.}
We build \MethodName\ on CronusVLA, a multi-frame post-trained OpenVLA policy with a DiT-style diffusion decoder~\cite{Kim2024,Li2025}. We replace its dense history conditioning with \ModuleName, which maps the history buffer to a fixed number of learned slots. Fine-tuning freezes the vision encoders and language model and optimizes the multimodal projector, LHC, and action decoder.

On Simpler-WidowX~\cite{li2024evaluating}, the LHC module distills $M=20$ history features into $K=6$ history slots. On LIBERO~\cite{liu2023libero}, the module uses $M=10$ history features and $K=3$ slots. Loss weights, optimization settings, diffusion sampling details, and hardware information are provided in the appendix.

\subsubsection{Evaluation Setup in SimplerEnv.}
We evaluate the four standard prepackaged SimplerEnv WidowX visual-matching tasks~\cite{li2024evaluating}: Spoon on Towel, Carrot on Plate, Stack Cube, and Eggplant in Basket. These probe manipulation robustness under a real-to-sim gap. Table~\ref{tab:simplerenv_results} reports mean success rates; the repeated-evaluation protocol, uncertainty estimates, simulator settings, and baseline provenance are given in Appendix~\ref{sec:appendix_simpler}.
The table includes local results and externally reported references with different training data and input modalities. Its layout does not imply a shared protocol for all rows; we focus our conclusions on the local CronusVLA--\MethodName\ comparison.

\subsubsection{Experimental Results on SimplerEnv.}
Across Simpler-WidowX tasks, \MethodName\ achieves a mean success rate of 69.8\%, compared with 58.7\% for CronusVLA (+11.1 pp). The gain supports improved manipulation robustness under recent-history conditioning; these tasks alone do not establish persistent memory capability.

\begin{table}[t]
  \caption{Mean success rate (\%) on Simpler-WidowX. Boldface marks lower-block best results; source and protocol differences are detailed in Appendix~\ref{sec:appendix_simpler}.}
  \label{tab:simplerenv_results}
  \centering
  \begin{tabular*}{\linewidth}{@{\extracolsep{\fill}}lccccc}
    \toprule
    Method & \shortstack{Spoon\\on Towel} & \shortstack{Carrot\\on Plate} & \shortstack{Stack\\Cube} & \shortstack{Eggplant\\in Basket} & \shortstack{Avg.\\Success} \\
    \midrule
    TraceVLA~\cite{Zheng2024} & 12.5 & 16.6 & 16.6 & 65.0 & 27.7 \\
    Octo-base~\cite{Ghosh2024} & 15.8 & 12.5 & 0.0 & 41.7 & 17.5 \\
    UniVLA~\cite{Bu2025} & 52.8 & 55.6 & 2.8 & 80.6 & 47.9 \\
    CogACT~\cite{Li2024a} & 75.0 & 50.0 & 16.7 & 79.2 & 55.2 \\
    Magma~\cite{Yang2025a} & 37.5 & 29.2 & 20.8 & 91.7 & 44.8 \\
    RT-1-X~\cite{ONeill2024} & 0.0 & 4.2 & 0.0 & 0.0 & 1.1 \\
    SpatialVLA~\cite{Qu2025} & 20.8 & 37.5 & 41.7 & 83.3 & 45.8 \\
    MemoryVLA~\cite{shi2025memoryvla} & 75.0 & 75.0 & 37.5 & 100.0 & 71.9 \\
    $\pi_0$~\cite{Black2024} & 46.7 & 38.7 & 42.7 & 39.3 & 41.8 \\
    ContextVLA~\cite{Jang2025} & 53.3 & 56.0 & 41.3 & 74.0 & 56.2 \\
    GR00T-N1.5~\cite{Bjorck2025} & 75.3 & 54.3 & 57.0 & 61.3 & 61.9 \\
    HAMLET~\cite{Koo2025HAMLET} & 75.0 & 62.5 & 16.7 & 100.0 & 63.5 \\
    \midrule
    OpenVLA~\cite{Kim2024} & 8.3 & 4.2 & 0.0 & 0.0 & 3.1 \\
    OpenVLA-OFT~\cite{Kim2025} & 12.5 & 4.2 & 4.2 & 95.8 & 29.2 \\
    CronusVLA~\cite{Li2025} & 66.7 & 51.4 & 22.2 & 94.4 & 58.7 \\
    \textbf{\MethodName} & \textbf{80.6} & \textbf{66.7} & \textbf{33.3} & \textbf{98.6} & \textbf{69.8} \\
    \bottomrule
  \end{tabular*}
\end{table}

The task-level gains over CronusVLA are +13.9/+15.3/+11.1/+4.2 pp for Spoon, Carrot, Stack, and Eggplant, respectively. Benefits are non-uniform, and these point estimates do not establish task-wise statistical significance. Qualitative occlusion overlays and slot-level temporal analyses are provided in the appendix.

\subsubsection{Evaluation Setup and Experimental Results on LIBERO.}
LIBERO consists of four suites: Spatial, Object, Goal, and Long, each containing 10 tasks that span spatial reasoning, object-centric manipulation, goal-conditioned adaptation, and long-horizon planning. Fine-tuning is performed from multi-frame post-training weights, and each suite is trained separately. The \ModuleName\ uses $K=3$ history slots and $M=10$ history features. CronusVLA and \MethodName\ use the same external-and-wrist RGB/no-state protocol, making CronusVLA the primary same-protocol baseline. Other methods are included as protocol-aware references; \(\pi_0\) and OpenVLA-OFT use wrist-view and robot-state inputs. Additional protocol details are provided in the appendix.

Under this setting, \MethodName\ achieves 97.2\% mean success, compared with 96.3\% for CronusVLA. On Long, the corresponding rates are 94.6\% and 94.0\%. The near-saturated scores support performance under this protocol, but do not alone isolate memory-dependent reasoning. Appendix~\ref{sec:appendix_libero} also includes the externally reported $\pi_{0.5}$ reference and protocol differences.

\begin{table}[t]
  \caption{Mean success rate (\%) on LIBERO. Repeated-evaluation and input protocols are reported in Appendix~\ref{sec:appendix_libero}.}
  \label{tab:libero_results}
  \centering
  \begin{tabular*}{\linewidth}{@{\extracolsep{\fill}}lccccc}
    \toprule
    Method & Spatial & Object & Goal & Long & Avg. Success \\
    \midrule
    Diffusion Policy~\cite{Chi2024DiffusionPolicy} & 78.3 & 92.5 & 68.3 & 50.5 & 72.4 \\
    OpenVLA~\cite{Kim2024} & 84.7 & 88.4 & 79.2 & 53.7 & 76.5 \\
    OpenVLA-OFT~\cite{Kim2025} & 97.6 & 98.4 & \textbf{97.9} & 94.5 & 97.1 \\
    $\pi_0$~\cite{Black2024} & 96.8 & 98.8 & 95.8 & 85.2 & 94.2 \\
    TraceVLA~\cite{Zheng2024} & 84.6 & 85.2 & 75.1 & 54.1 & 74.8 \\
    SpatialVLA~\cite{Qu2025} & 88.2 & 89.9 & 78.6 & 55.5 & 78.1 \\
    Dita~\cite{Hou2025Dita} & 84.2 & 96.3 & 85.4 & 63.8 & 82.4 \\
    CronusVLA~\cite{Li2025} & 97.1 & 98.1 & 96.0 & 94.0 & 96.3 \\
    \textbf{\MethodName} & \textbf{97.8} & \textbf{99.1} & 97.4 & \textbf{94.6} & \textbf{97.2} \\
    \bottomrule
  \end{tabular*}
\end{table}

\subsubsection{Memory-dependent Evaluation on RMBench.}
\label{sec:rmbench}
Simpler primarily measures manipulation robustness, while LIBERO is close to saturation; neither alone isolates the need for historical observations. We therefore evaluate on RMBench~\cite{chen2026rmbench}, whose $M(1)$ and $M(n)$ tasks require retaining one or multiple task-relevant past observations, respectively. These categories count required observations, not their temporal distance. Each of the nine tasks uses 50 demonstrations and 100 evaluation rollouts.

The primary controlled comparison uses dense CronusVLA and \MethodName\ with the same rolling FIFO of the most recent 20 historical features. The current-frame feature is processed separately and identically. Initialization, training data, training steps, seed, action horizon, and evaluation protocol are matched; \MethodName\ condenses the history to $K=6$ slots, transferring the Simpler configuration without RMBench-specific retuning. We also report a current-frame-only OpenVLA reference; the dense-versus-condensed pair isolates the history-conditioning change.

\begin{table}[t]
  \centering
  \caption{RMBench success rate (\%). $M(1)$/$M(n)$ denote memory-demand categories, distinct from history length $M$. Overall averages all nine tasks.}
  \label{tab:rmbench_results}
  \begin{tabular*}{\linewidth}{@{\extracolsep{\fill}}lccc}
    \toprule
    Method & $M(1)$ & $M(n)$ & Overall \\
    \midrule
    OpenVLA (current frame) & 5.4 & 3.8 & 4.7 \\
    CronusVLA (dense, $M=20$) & 18.2 & 9.5 & 14.3 \\
    \textbf{\MethodName} ($M=20$, $K=6$) & \textbf{28.8} & \textbf{12.5} & \textbf{21.6} \\
    \bottomrule
  \end{tabular*}
\end{table}

\MethodName\ improves overall success from 14.3 to 21.6 (+7.2 percentage points, pp), with larger gains on $M(1)$ (+10.6 pp) than on $M(n)$ (+3.0 pp). All three methods fail on Press Button. These results support selective condensation within a bounded recent window; low $M(n)$ success also exposes its limits. Once an event leaves the FIFO, our history mechanism has no explicit retrieval path, persistent semantic task state, or high-level planner. Appendix~\ref{sec:appendix_rmbench} provides every task result and the separately reported Mem-0 reference.

\subsection{Real-world Experimental Results}

\subsubsection{Evaluation Setup on the Galaxea R1 Lite Platform.}
Real-world evaluation uses a Galaxea R1 Lite robot and a third-person RGB camera mounted at the top center of its body. \MethodName\ uses $M=20$ historical features and $K=6$ slots. The two simple tasks are Put Apple on Plate and Open then Close Drawer. The three multi-stage tasks are Pick and Place Twice (transfer the apple to the other plate and return it), Swap Cubes, and Put Objects in Drawer (place the cube and apple inside and close the drawer). Each task uses 50 training demonstrations and 50 evaluation trials per method. We compare OpenVLA-OFT, $\pi_0$, CronusVLA, and \MethodName; Appendix~\ref{sec:appendix_realworld_discussion} provides uncertainty estimates and task details.

\subsubsection{Experimental Results on the Galaxea R1 Lite Platform.}
\MethodName\ achieves 52.8\% mean final success versus 45.2\% for CronusVLA (+7.6 pp). Across the three multi-stage tasks, mean final success improves from 37.3\% to 46.7\% (+9.3 pp, computed before rounding). Final success rises from 40\% to 50\% on Pick and Place Twice, 50\% to 62\% on Swap Cubes, and 22\% to 28\% on Put Objects in Drawer. Gains vary across tasks and stages; we report every stage in Table~\ref{tab:realworld_results}. Representative rollouts are shown in \cref{fig:realworld_process}, with failure cases in the appendix.

\begin{figure}[t]
  \centering
  \includegraphics[width=1.0\linewidth]{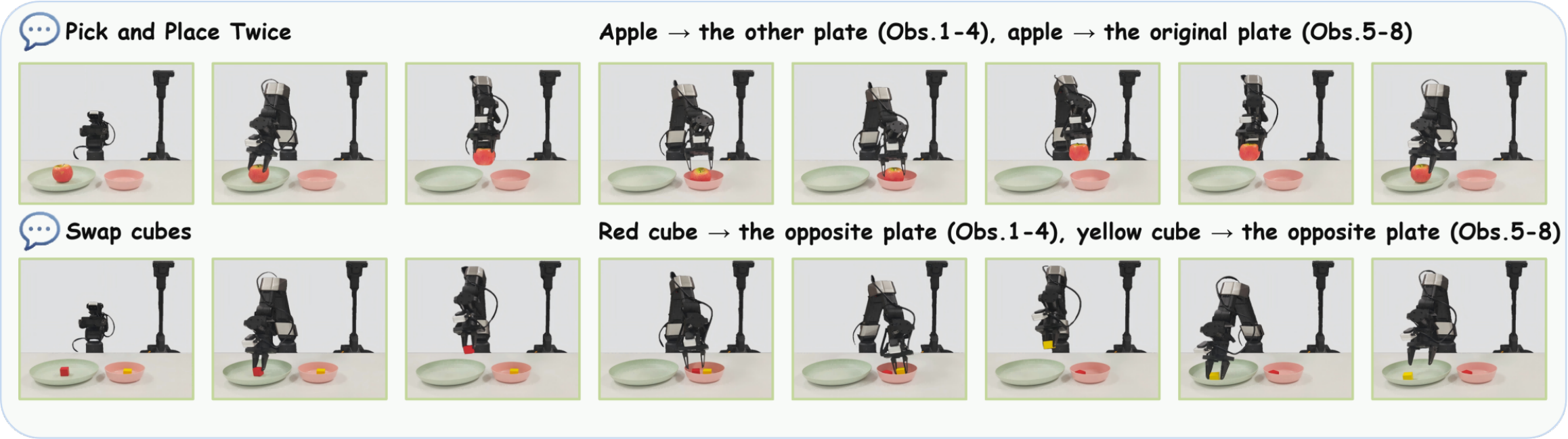}
  \caption{Real-world execution snapshots on the Galaxea R1 Lite platform. The figure shows representative stage-wise trajectories for two long-horizon tasks: Pick and Place Twice and Swap Cubes.}
  \label{fig:realworld_process}
\end{figure}

\begin{table}[t]
  \caption{Success rate (\%) over 50 trials per task and method on Galaxea R1 Lite. Intermediate stages precede final success; Avg. averages final success only. Put Objects in Drawer requires both insertions and drawer closure.}
  \label{tab:realworld_results}
  \centering
  \scriptsize
  \setlength{\tabcolsep}{1.5pt}
  \begin{tabular*}{\linewidth}{@{\extracolsep{\fill}}lcccc|ccccccc}
    \toprule
    Method & \multicolumn{2}{c}{\shortstack{Put Apple\\on Plate}} & \multicolumn{2}{c|}{\shortstack{Open then\\Close Drawer}} & \multicolumn{2}{c}{\shortstack{Pick and\\Place Twice}} & \multicolumn{2}{c}{\shortstack{Swap\\Cubes}} & \multicolumn{2}{c}{\shortstack{Put Objects\\in Drawer}} & \shortstack{Avg.\\Success} \\
    \cmidrule(lr){2-3} \cmidrule(lr){4-5} \cmidrule(lr){6-7} \cmidrule(lr){8-9} \cmidrule(lr){10-11}
    & Grasp & Success & Open & Success & PnP Once & Success & First Cube & Success & Obj.~1 Placed & Success & \\
    \midrule
    OpenVLA-OFT~\cite{Kim2025} & 52.0 & 48.0 & 56.0 & 44.0 & 48.0 & 16.0 & 48.0 & 8.0 & 12.0 & 0.0 & 23.2 \\
    $\pi_0$~\cite{Black2024} & \textbf{72.0} & \textbf{72.0} & \textbf{68.0} & \textbf{60.0} & \textbf{56.0} & 24.0 & 64.0 & 12.0 & 24.0 & 16.0 & 36.8 \\
    CronusVLA~\cite{Li2025} & 68.0 & 66.0 & 52.0 & 48.0 & 46.0 & 40.0 & 58.0 & 50.0 & 28.0 & 22.0 & 45.2 \\
    \textbf{\MethodName} & \textbf{72.0} & 70.0 & 56.0 & 54.0 & 52.0 & \textbf{50.0} & \textbf{66.0} & \textbf{62.0} & \textbf{32.0} & \textbf{28.0} & \textbf{52.8} \\
    \bottomrule
  \end{tabular*}
\end{table}

\subsection{Ablation Study}
All Simpler ablations below use the original 24-rollout-per-task protocol, including its reported 58.4/70.8 baseline/final averages. They are separate from the repeated-checkpoint evaluation in Table~\ref{tab:simplerenv_results}; its uncertainty estimates do not apply to these ablation variants.

\subsubsection{Component Ablations.}
\begin{figure}[t]
  \centering
  \setlength{\tabcolsep}{3pt}
  \begin{tabular}{cc}
    \includegraphics[height=0.33\linewidth]{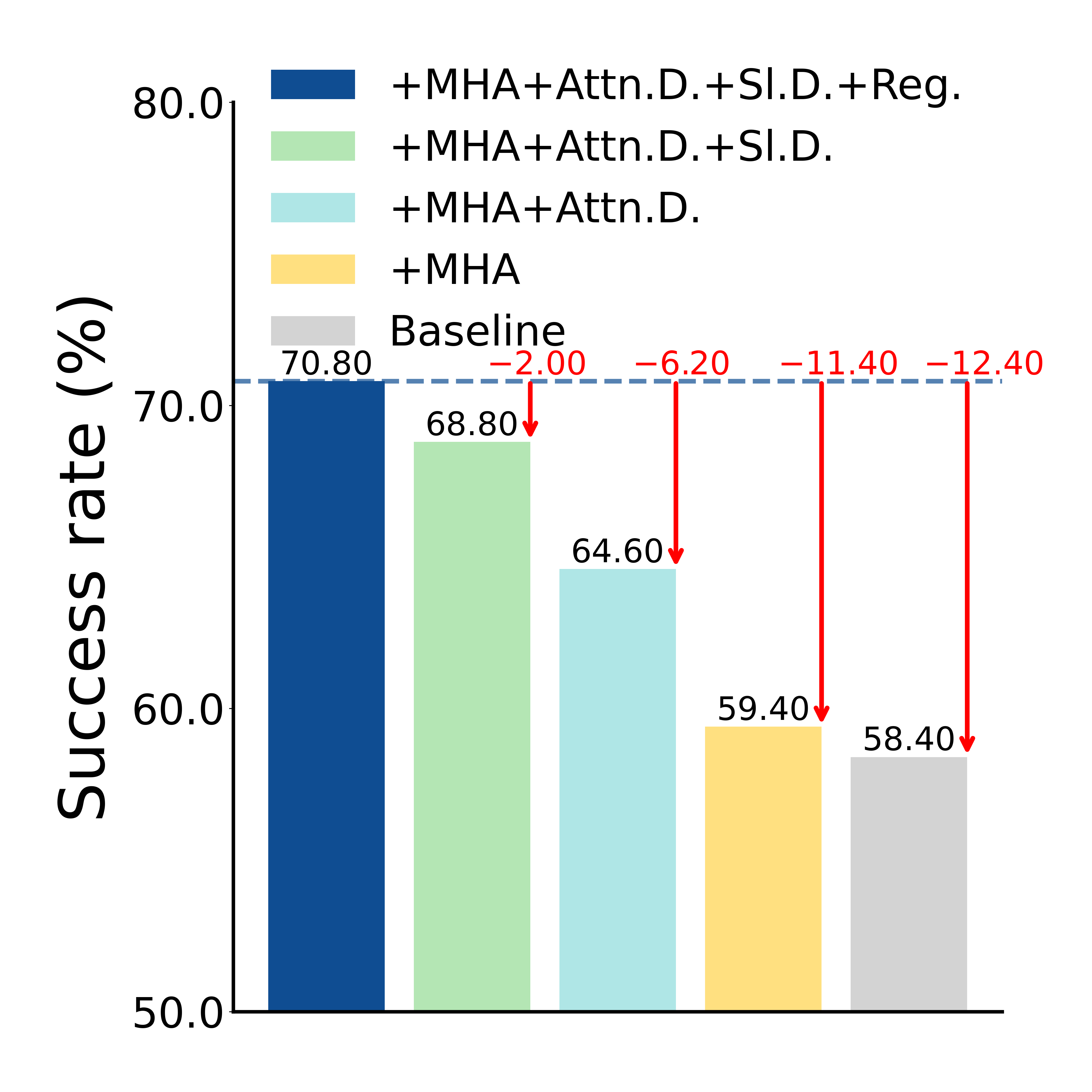} &
    \includegraphics[height=0.30\linewidth]{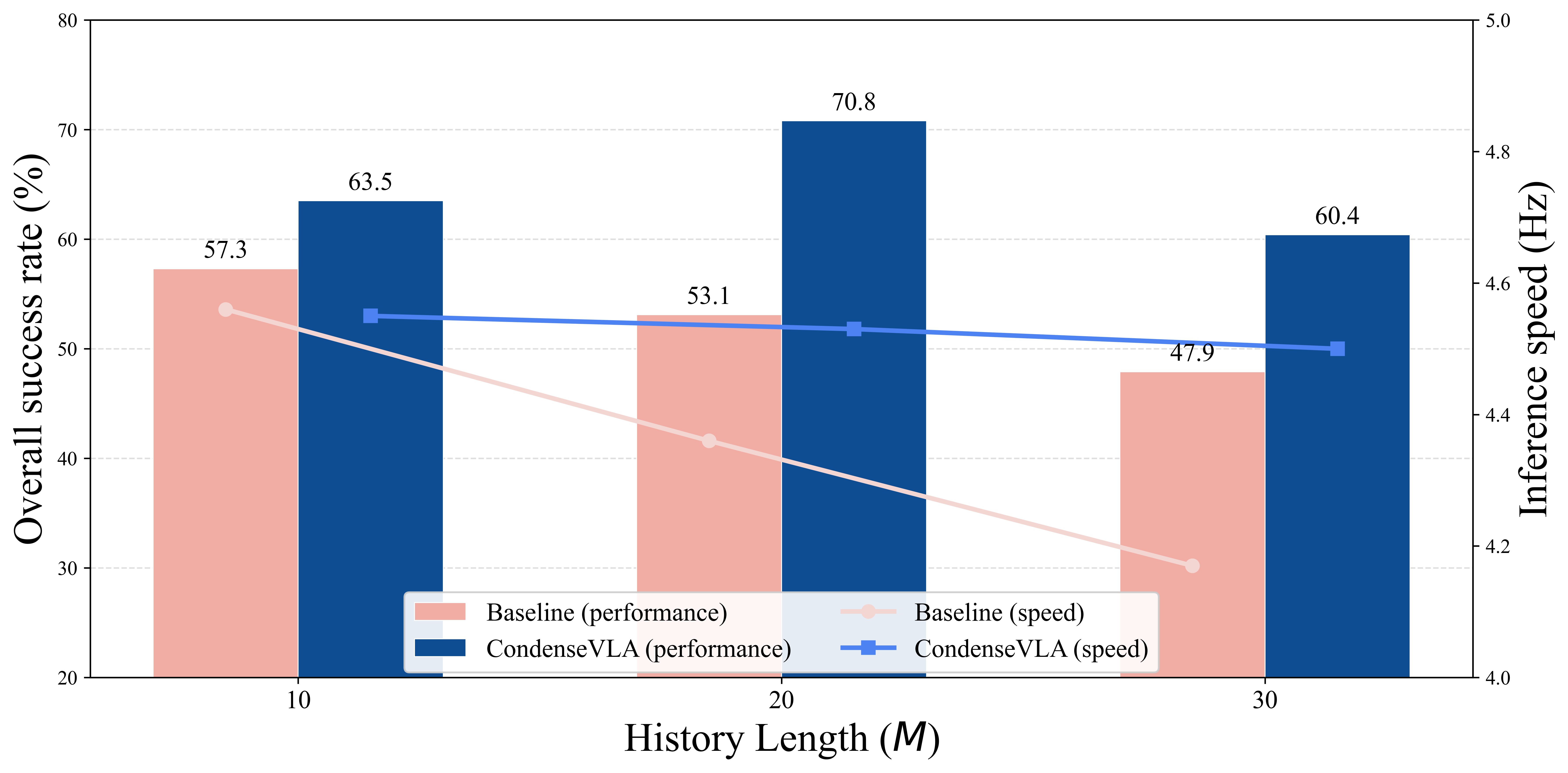} \\
    (a) \ModuleShort\ design ablation & (b) History-length ablation
  \end{tabular}
  \caption{(a) Component ablations under the original 24-rollout protocol. The final configuration (labeled ``+Reg.'') replaces all-pairs with middle-only slot decorrelation. (b) Success rate under the same ablation protocol and inference speed at fixed $K=6$; Baseline denotes dense CronusVLA. Speed (Hz) is the reciprocal of measured policy-forward latency, using an A100-SXM4-80GB, batch size 1, and 10-step DDIM. Appendix~\ref{sec:appendix_latency} gives the profiling protocol and component breakdown.}
  \label{fig:ablation_selector_design}
  \label{fig:ablation_frame_number_twin}
\end{figure}

We compare dense-history conditioning, MHA-based condensation, attention diversity, and slot decorrelation. Fig.~\ref{fig:ablation_selector_design}(a) ends by replacing all-pairs slot decorrelation with middle-only decorrelation; this changes the pair mask, not the number of auxiliary losses.

\subsubsection{Effect of History Length and Condensation Budget.}
At fixed $K=6$, \MethodName\ reaches 70.8\% success at $M=20$ under the ablation protocol. Fig.~\ref{fig:ablation_selector_design}(b) reports success and inference speed as history grows. Measured policy latency is 0.25\% higher than dense conditioning at $M=10$, but 3.8\% and 7.4\% lower at $M=20$ and $M=30$. The fixed decoder-conditioning budget limits decoder cost, while LHC cost still grows with $M$; the results do not establish a universal end-to-end speedup.

Table~\ref{tab:mk_scaling_matrix} shows an interaction between $M$ and $K$: $K=6$ is best at $M=20$, but $K=10$ is best at $M=30$. We selected $M=20,K=6$ on Simpler and transferred it unchanged to RMBench. No separate held-out hyperparameter split was used. These results support joint selection of history length and slot budget for the target setting.

\begin{table}[t]
  \caption{K/M scaling on Simpler-WidowX. Each cell reports success rate (\%).}
  \label{tab:mk_scaling_matrix}
  \centering
  \small
  \setlength{\tabcolsep}{8pt}
  \begin{tabular*}{\linewidth}{@{\extracolsep{\fill}}lccc}
    \toprule
    $M \backslash K$ & $K=3$ & $K=6$ & $K=10$ \\
    \midrule
    $M=10$ & 59.4 & 63.5 & 56.2 \\
    $M=20$ & 57.3 & \textbf{70.8} & 62.5 \\
    $M=30$ & 53.1 & 60.4 & 64.6 \\
    \bottomrule
  \end{tabular*}
\end{table}

\subsubsection{Comparisons with Fixed History Selection.}
We compare learned condensation with Static-Index, Static-Weight, and Strided History Sampling. The two endpoint templates replace LHC at inference without retraining, bypassing its projections and time embeddings. Strided sampling selects six positions $[0,4,8,11,15,19]$ from the $M=20$ history using zero-based indexing. Appendix details distinguish the endpoint interventions from independently trained baselines.

The tested alternatives score below learned condensation in \cref{tab:template_ablation}. Matching $M$ and $K$ controls history and output budgets, but these comparisons do not isolate adaptive attention from learned projections, capacity, or the conditioning-distribution change caused by inference-time replacement.

\begin{table}[!htbp]
\centering
\caption{History-selection comparisons on Simpler-WidowX; static templates are inference-time replacements. Gap is variant SR minus learned SR, in percentage points (pp).}
\label{tab:template_ablation}
\small
\setlength{\tabcolsep}{4pt}
\begin{tabular*}{\linewidth}{@{\extracolsep{\fill}}lcccc}
\hline
Method & $M$ & $K$ & SR (\%) & Gap (pp) \\
\hline
\textbf{\MethodName\ (learned) } & 20 & 6 & \textbf{70.8} & 0.0 \\
Static-Index Selection & 20 & 6 & 55.2 & -15.6 \\
Static-Weight Selection & 20 & 6 & 54.2 & -16.6 \\
Strided History Sampling & 20 & 6 & 59.4 & -11.4 \\
\hline
\end{tabular*}
\end{table}

Appendix attention visualizations show different temporal coverage across slots and tasks. These descriptive patterns are consistent with adaptive history allocation, but do not assign fixed temporal roles to slot indices or identify a sole cause of the performance gain.

\paragraph{Limitations.}
Our FIFO cannot explicitly retrieve events outside its window, and RMBench $M(n)$ success remains low. Results are sensitive to $M$ and $K$, selected without a separate held-out tuning split, so they do not establish a robust default across tasks. Fixed-checkpoint uncertainty estimates exclude training randomness. Evaluation uses one VLA backbone family and one physical robot platform, and condensation cannot recover evidence absent from frozen visual features. Added LHC capacity also prevents attributing all gains to temporal selection alone. Appendix~\ref{sec:appendix_limitations} discusses these boundaries.

\section{Conclusion}
We present \MethodName, a deployment-oriented history-conditioning mechanism that condenses recent observation histories into fixed slot tokens for diffusion-based VLA policies. Across benchmarks, it improves Simpler-WidowX success from 58.7\% to 69.8\%, achieves 97.2\% on LIBERO, reaches 52.8\% on real-world tasks, and achieves 21.6\% on RMBench. These results support a practical principle for multi-frame VLA design: temporal context should be condensed and structured, not merely enlarged.

% Our study is limited to a specific diffusion-VLA backbone and a modest set of simulation/real-robot settings, and slot semantics remain only weakly interpretable. Future work will prioritize broader cross-embodiment validation and more interpretable, adaptively allocated memory slots.

\clearpage

\begin{ack}
Shuo Yang is supported by the Shenzhen Fundamental Research Program (JCYJ20250604145514018), Guangdong Basic and Applied Basic Research Foundation (General Program, No. 2026A1515011557), Guangdong S\&T Programme (Grant No.~2025B0101130003) and the NSFC Young Scientists Fund (No. 62506096).

The authors declare no competing interests.
\end{ack}

{\small
\bibliographystyle{plainnat}
\bibliography{main}

\begin{thebibliography}{58}
\providecommand{\natexlab}[1]{#1}
\providecommand{\url}[1]{\texttt{#1}}
\expandafter\ifx\csname urlstyle\endcsname\relax
  \providecommand{\doi}[1]{doi: #1}\else
  \providecommand{\doi}{doi: \begingroup \urlstyle{rm}\Url}\fi

\bibitem[Bjorck et~al.(2025)Bjorck, Casta{\~n}eda, Cherniadev, Da, Ding, Fan,
  Fang, Fox, Hu, Huang, et~al.]{Bjorck2025}
Johan Bjorck, Fernando Casta{\~n}eda, Nikita Cherniadev, Xingye Da, Runyu Ding,
  Linxi Fan, Yu~Fang, Dieter Fox, Fengyuan Hu, Spencer Huang, et~al.
\newblock Gr00t n1: An open foundation model for generalist humanoid robots.
\newblock \emph{arXiv preprint arXiv:2503.14734}, 2025.

\bibitem[Black et~al.(2025)Black, Brown, Driess, Esmail, Equi, Finn, Fusai,
  Groom, Hausman, Ichter, et~al.]{Black2024}
Kevin Black, Noah Brown, Danny Driess, Adnan Esmail, Michael Equi, Chelsea
  Finn, Niccolo Fusai, Lachy Groom, Karol Hausman, Brian Ichter, et~al.
\newblock $\pi$0: A vision-language-action flow model for general robot
  control.
\newblock \emph{RSS}, 2025.

\bibitem[Brohan et~al.(2023)Brohan, Brown, Carbajal, Chebotar, Dabis, Finn,
  Gopalakrishnan, Hausman, Herzog, Hsu, et~al.]{brohan2023rt1}
Anthony Brohan, Noah Brown, Justice Carbajal, Yevgen Chebotar, Joseph Dabis,
  Chelsea Finn, Keerthana Gopalakrishnan, Karol Hausman, Alex Herzog, Jasmine
  Hsu, et~al.
\newblock Rt-1: Robotics transformer for real-world control at scale.
\newblock \emph{RSS}, 2023.

\bibitem[Bu et~al.(2025)Bu, Yang, Cai, Gao, Ren, Yao, Luo, and Li]{Bu2025}
Qingwen Bu, Yanting Yang, Jisong Cai, Shenyuan Gao, Guanghui Ren, Maoqing Yao,
  Ping Luo, and Hongyang Li.
\newblock Univla: Learning to act anywhere with task-centric latent actions.
\newblock \emph{RSS}, 2025.

\bibitem[Chen et~al.(2026{\natexlab{a}})Chen, Fang, Wang, Wang, and
  Lu]{Chen2025}
Jingjing Chen, Hongjie Fang, Chenxi Wang, Shiquan Wang, and Cewu Lu.
\newblock History-aware visuomotor policy learning via point tracking.
\newblock \emph{ICRA}, 2026{\natexlab{a}}.

\bibitem[Chen et~al.(2026{\natexlab{b}})Chen, Wang, Li, Qin, Shi, Li, Hu,
  Zhang, Wang, Chen, et~al.]{chen2026rmbench}
Tianxing Chen, Yuran Wang, Mingleyang Li, Yan Qin, Hao Shi, Zixuan Li, Yifan
  Hu, Yingsheng Zhang, Kaixuan Wang, Yue Chen, et~al.
\newblock Rmbench: Memory-dependent robotic manipulation benchmark with
  insights into policy design.
\newblock \emph{arXiv preprint arXiv:2603.01229}, 2026{\natexlab{b}}.

\bibitem[Chi et~al.(2025)Chi, Xu, Feng, Cousineau, Du, Burchfiel, Tedrake, and
  Song]{Chi2024DiffusionPolicy}
Cheng Chi, Zhenjia Xu, Siyuan Feng, Eric Cousineau, Yilun Du, Benjamin
  Burchfiel, Russ Tedrake, and Shuran Song.
\newblock Diffusion policy: Visuomotor policy learning via action diffusion.
\newblock \emph{IJRR}, 2025.

\bibitem[Choi et~al.(2024)Choi, Lee, Chu, Choi, and Kim]{Choi2024}
Joonmyung Choi, Sanghyeok Lee, Jaewon Chu, Minhyuk Choi, and Hyunwoo~J Kim.
\newblock vid-tldr: Training free token merging for light-weight video
  transformer.
\newblock \emph{CVPR}, 2024.

\bibitem[Dai et~al.(2026)Dai, Lan, Zhou, Zhao, Su, Tong, Guan, and
  Yang]{dai2026conla}
Weisheng Dai, Kai Lan, Jianyi Zhou, Bo~Zhao, Xiu Su, Junwen Tong, Weili Guan,
  and Shuo Yang.
\newblock Conla: Contrastive latent action learning from human videos for
  robotic manipulation.
\newblock \emph{arXiv preprint arXiv:2602.00557}, 2026.

\bibitem[Driess et~al.(2023)Driess, Xia, Sajjadi, Lynch, Chowdhery, Ichter,
  Wahid, Tompson, Vuong, Yu, et~al.]{Driess2023PaLME}
Danny Driess, Fei Xia, Mehdi~SM Sajjadi, Corey Lynch, Aakanksha Chowdhery,
  Brian Ichter, Ayzaan Wahid, Jonathan Tompson, Quan Vuong, Tianhe Yu, et~al.
\newblock Palm-e: An embodied multimodal language model.
\newblock \emph{ICML}, 2023.

\bibitem[Ghosh et~al.(2024)Ghosh, Walke, Yang, Ma, Black, et~al.]{Ghosh2024}
Dibya Ghosh, Homer Walke, Yizhou Yang, Yao Ma, Kevin Black, et~al.
\newblock Octo: An open-source generalist robot policy.
\newblock \emph{arXiv preprint arXiv:2405.12213}, 2024.

\bibitem[Guo et~al.(2025)Guo, Yang, Huang, Long, Xia, Su, Zhao, Xie, and
  Nie]{guo2026utilgen}
Jiyu Guo, Shuo Yang, Yiming Huang, Yancheng Long, Xiaobo Xia, Xiu Su, Bo~Zhao,
  Zeke Xie, and Liqiang Nie.
\newblock Utilgen: Utility-centric generative data augmentation with dual-level
  task adaptation.
\newblock \emph{NeurIPS}, 38:\penalty0 32474--32500, 2025.

\bibitem[Hanyu et~al.(2025)Hanyu, Chung, Le, Nguyen, Ikebe, Gunderman, Minh,
  Vo, Kieu, Yamazaki, et~al.]{Hanyu2025}
Taisei Hanyu, Nhat Chung, Huy Le, Toan Nguyen, Yuki Ikebe, Anthony Gunderman,
  Duy Nguyen~Ho Minh, Khoa Vo, Tung Kieu, Kashu Yamazaki, et~al.
\newblock Slotvla: Towards modeling of object-relation representations in
  robotic manipulation.
\newblock \emph{arXiv preprint arXiv:2511.06754}, 2025.

\bibitem[Hou et~al.(2025)Hou, Zhang, Xiong, Duan, Pu, Tong, Zhao, Zhu, Qiao,
  Dai, and Chen]{Hou2025Dita}
Zhi Hou, Tianyi Zhang, Yuwen Xiong, Haonan Duan, Hengjun Pu, Ronglei Tong,
  Chengyang Zhao, Xizhou Zhu, Yu~Qiao, Jifeng Dai, and Yuntao Chen.
\newblock Dita: Scaling diffusion transformer for generalist
  vision-language-action policy.
\newblock \emph{ICCV}, 2025.

\bibitem[Huang et~al.(2025)Huang, Wang, Lin, Hu, Wen, and
  Gao]{huang2025tactile}
Jialei Huang, Shuo Wang, Fanqi Lin, Yihang Hu, Chuan Wen, and Yang Gao.
\newblock Tactile-vla: unlocking vision-language-action model's physical
  knowledge for tactile generalization.
\newblock \emph{arXiv preprint arXiv:2507.09160}, 2025.

\bibitem[Intelligence et~al.(2025)Intelligence, Black, Brown, Darpinian,
  Dhabalia, Driess, Esmail, Equi, Finn, Fusai, et~al.]{Black2025pi05}
Physical Intelligence, Kevin Black, Noah Brown, James Darpinian, Karan
  Dhabalia, Danny Driess, Adnan Esmail, Michael Equi, Chelsea Finn, Niccolo
  Fusai, et~al.
\newblock $\pi_{0.5}$: a vision-language-action model with open-world
  generalization.
\newblock \emph{CoRL}, 2025.

\bibitem[Jang et~al.(2025)Jang, Yu, Kwon, Jeon, Seo, and Shin]{Jang2025}
Huiwon Jang, Sihyun Yu, Heeseung Kwon, Hojin Jeon, Younggyo Seo, and Jinwoo
  Shin.
\newblock Contextvla: Vision-language-action model with amortized multi-frame
  context.
\newblock \emph{arXiv preprint arXiv:2510.04246}, 2025.

\bibitem[Kim et~al.(2024)Kim, Pertsch, Karamcheti, Xiao, Balakrishna, Nair,
  Rafailov, Foster, Lam, Sanketi, et~al.]{Kim2024}
Moo~Jin Kim, Karl Pertsch, Siddharth Karamcheti, Ted Xiao, Ashwin Balakrishna,
  Suraj Nair, Rafael Rafailov, Ethan Foster, Grace Lam, Pannag Sanketi, et~al.
\newblock Openvla: An open-source vision-language-action model.
\newblock \emph{arXiv preprint arXiv:2406.09246}, 2024.

\bibitem[Kim et~al.(2025)Kim, Finn, and Liang]{Kim2025}
Moo~Jin Kim, Chelsea Finn, and Percy Liang.
\newblock Fine-tuning vision-language-action models: Optimizing speed and
  success.
\newblock \emph{RSS}, 2025.

\bibitem[Koo et~al.(2026)Koo, Choi, Kim, Lee, Kim, Seo, and
  Shin]{Koo2025HAMLET}
Myungkyu Koo, Daewon Choi, Taeyoung Kim, Kyungmin Lee, Changyeon Kim, Younggyo
  Seo, and Jinwoo Shin.
\newblock Hamlet: Switch your vision-language-action model into a history-aware
  policy.
\newblock \emph{ICLR}, 2026.

\bibitem[Lee et~al.(2025)Lee, Ekpo, Liu, Huang, Shrivastava, and
  Huang]{lee2025imagine}
Seungjae Lee, Daniel Ekpo, Haowen Liu, Furong Huang, Abhinav Shrivastava, and
  Jia-Bin Huang.
\newblock Imagine, verify, execute: Memory-guided agentic exploration with
  vision-language models.
\newblock \emph{CoRL}, 2025.

\bibitem[Li et~al.(2025{\natexlab{a}})Li, Wu, Yan, Liu, Zeng, Li, and
  Zha]{li2025reflection}
Baicheng Li, Dong Wu, Zike Yan, Xinchen Liu, Zecui Zeng, Lusong Li, and Hongbin
  Zha.
\newblock Reflection-based task adaptation for self-improving vla.
\newblock \emph{arXiv preprint arXiv:2510.12710}, 2025{\natexlab{a}}.

\bibitem[Li et~al.(2025{\natexlab{b}})Li, Yang, Chen, Chen, Yang, Tian, Wang,
  Wang, Lin, Zhao, et~al.]{Li2025}
Hao Li, Shuai Yang, Yilun Chen, Xinyi Chen, Xiaoda Yang, Yang Tian, Hanqing
  Wang, Tai Wang, Dahua Lin, Feng Zhao, et~al.
\newblock Cronusvla: Towards efficient and robust manipulation via multi-frame
  vision-language-action modeling.
\newblock \emph{arXiv preprint arXiv:2506.19816}, 2025{\natexlab{b}}.

\bibitem[Li et~al.(2024{\natexlab{a}})Li, Liang, Wang, Luo, Chen, Liao, Wei,
  Deng, Xu, Zhang, et~al.]{Li2024a}
Qixiu Li, Yaobo Liang, Zeyu Wang, Lin Luo, Xi~Chen, Mozheng Liao, Fangyun Wei,
  Yu~Deng, Sicheng Xu, Yizhong Zhang, et~al.
\newblock {CogACT}: A foundational vision-language-action model for synergizing
  cognition and action in robotic manipulation.
\newblock \emph{arXiv preprint arXiv:2411.19650}, 2024{\natexlab{a}}.

\bibitem[Li et~al.(2025{\natexlab{c}})Li, Zhang, Shao, He, and
  Nie]{li2025cogvla}
Wei Li, Renshan Zhang, Rui Shao, Jie He, and Liqiang Nie.
\newblock Cogvla: Cognition-aligned vision-language-action model via
  instruction-driven routing \& sparsification.
\newblock \emph{NeurIPS}, 2025{\natexlab{c}}.

\bibitem[Li et~al.(2026)Li, Zhang, Shao, Fang, Zhou, Tian, and
  Nie]{li2026SemanticVLA}
Wei Li, Renshan Zhang, Rui Shao, Zhijian Fang, Kaiwen Zhou, Zhuotao Tian, and
  Liqiang Nie.
\newblock Semanticvla: Semantic-aligned sparsification and enhancement for
  efficient robotic manipulation.
\newblock \emph{AAAI}, 2026.

\bibitem[Li et~al.(2024{\natexlab{b}})Li, Li, Liu, Wang, Liu, Kang, Ma, Kong,
  Zhang, and Liu]{Li2024c}
Xinghang Li, Peiyan Li, Minghuan Liu, Dong Wang, Jirong Liu, Bingyi Kang, Xiao
  Ma, Tao Kong, Hanbo Zhang, and Huaping Liu.
\newblock What matters in building vision-language-action models.
\newblock \emph{NeurIPS}, 2024{\natexlab{b}}.

\bibitem[Li et~al.(2024{\natexlab{c}})Li, Hsu, Gu, Pertsch, Mees, Walke, Fu,
  Lunawat, Sieh, Kirmani, et~al.]{li2024evaluating}
Xuanlin Li, Kyle Hsu, Jiayuan Gu, Karl Pertsch, Oier Mees, Homer~Rich Walke,
  Chuyuan Fu, Ishikaa Lunawat, Isabel Sieh, Sean Kirmani, et~al.
\newblock Evaluating real-world robot manipulation policies in simulation.
\newblock \emph{CoRL}, 2024{\natexlab{c}}.

\bibitem[Li et~al.(2025{\natexlab{d}})Li, Mao, Yuan, Sim, Eisner, and
  Held]{li2025learn}
Yishu Li, Xinyi Mao, Ying Yuan, Kyutae Sim, Ben Eisner, and David Held.
\newblock Learn from what we have: History-aware verifier that reasons about
  past interactions online.
\newblock \emph{CoRL}, 2025{\natexlab{d}}.

\bibitem[Li et~al.(2025{\natexlab{e}})Li, Chen, Zhou, Li, Zhang, and
  Zhao]{li2025qdepth}
Yixuan Li, Yuhui Chen, Mingcai Zhou, Haoran Li, Zhengtao Zhang, and Dongbin
  Zhao.
\newblock Qdepth-vla: quantized depth prediction as auxiliary supervision for
  vision-language-action models.
\newblock \emph{arXiv preprint arXiv:2510.14836}, 2025{\natexlab{e}}.

\bibitem[Liu et~al.(2023)Liu, Zhu, Gao, Feng, Liu, Zhu, and
  Stone]{liu2023libero}
Bo~Liu, Yifeng Zhu, Chongkai Gao, Yihao Feng, Qiang Liu, Yuke Zhu, and Peter
  Stone.
\newblock Libero: Benchmarking knowledge transfer for lifelong robot learning.
\newblock \emph{NeurIPS}, 2023.

\bibitem[Liu et~al.(2026{\natexlab{a}})Liu, Miao, Chen, Yin, Zhu, Pang, Shou,
  Yan, Wang, and Guo]{liu2026worldactionlearninginteractioncentric}
Zhiming Liu, Yikun Miao, Ying Chen, Hongrui Yin, Fangqi Zhu, Xiaoyi Pang,
  Quanxin Shou, Zhengyang Yan, Haodong Wang, and Song Guo.
\newblock World action learning via interaction-centric spectral latent
  guidance.
\newblock \emph{arXiv preprint arXiv:2610.03607}, 2026{\natexlab{a}}.

\bibitem[Liu et~al.(2026{\natexlab{b}})Liu, Wei, Feng, Su, Xia, Guan, Xie, and
  Yang]{liu2026all}
Zhiming Liu, Yujie Wei, Lei Feng, Xiu Su, Xiaobo Xia, Weili Guan, Zeke Xie, and
  Shuo Yang.
\newblock Do all individual layers help? an empirical study of task-interfering
  layers in vision-language models.
\newblock \emph{arXiv preprint arXiv:2602.01167}, 2026{\natexlab{b}}.

\bibitem[O’Neill et~al.(2024)O’Neill, Rehman, Maddukuri, Gupta, Padalkar,
  Lee, Pooley, Gupta, Mandlekar, Jain, et~al.]{ONeill2024}
Abby O’Neill, Abdul Rehman, Abhiram Maddukuri, Abhishek Gupta, Abhishek
  Padalkar, Abraham Lee, Acorn Pooley, Agrim Gupta, Ajay Mandlekar, Ajinkya
  Jain, et~al.
\newblock Open x-embodiment: Robotic learning datasets and rt-x models: Open
  x-embodiment collaboration 0.
\newblock \emph{ICRA}, 2024.

\bibitem[Peebles and Xie(2023)]{Peebles2023DiT}
William Peebles and Saining Xie.
\newblock Scalable diffusion models with transformers.
\newblock \emph{ICCV}, 2023.

\bibitem[Pertsch et~al.(2025)Pertsch, Stachowicz, Ichter, Driess, Nair, Vuong,
  Mees, Finn, and Levine]{Pertsch2025}
Karl Pertsch, Kyle Stachowicz, Brian Ichter, Danny Driess, Suraj Nair, Quan
  Vuong, Oier Mees, Chelsea Finn, and Sergey Levine.
\newblock Fast: Efficient action tokenization for vision-language-action
  models.
\newblock \emph{arXiv preprint arXiv:2501.09747}, 2025.

\bibitem[Qian et~al.(2026)Qian, Wang, Hu, Zhang, and Zhang]{qian2026eida}
Yiwei Qian, Shanze Wang, Qingyuan Hu, Xinming Zhang, and Wei Zhang.
\newblock {EIDA}: Execution-interface dynamics adaptation for
  real-to-sim-to-real robot navigation.
\newblock \emph{arXiv preprint arXiv:2610.01219}, 2026.

\bibitem[Qu et~al.(2025)Qu, Song, Chen, Yao, Ye, Ding, Wang, Gu, Zhao, Wang,
  and Li]{Qu2025}
Delin Qu, Haoming Song, Qizhi Chen, Yuanqi Yao, Xinyi Ye, Yan Ding, Zhigang
  Wang, Jiayuan Gu, Bin Zhao, Dong Wang, and Xuelong Li.
\newblock Spatialvla: Exploring spatial representations for
  visual-language-action model.
\newblock \emph{RSS}, 2025.

\bibitem[Shi et~al.(2026)Shi, Xie, Liu, Sun, Liu, Wang, Zhou, Fan, Zhang, and
  Huang]{shi2025memoryvla}
Hao Shi, Bin Xie, Yingfei Liu, Lin Sun, Fengrong Liu, Tiancai Wang, Erjin Zhou,
  Haoqiang Fan, Xiangyu Zhang, and Gao Huang.
\newblock Memoryvla: Perceptual-cognitive memory in vision-language-action
  models for robotic manipulation.
\newblock \emph{ICLR}, 2026.

\bibitem[Sridhar et~al.(2026)Sridhar, Pan, Sharma, and Finn]{Sridhar2026MemER}
Ajay Sridhar, Jennifer Pan, Satvik Sharma, and Chelsea Finn.
\newblock Scaling up memory for robotic control via experience retrieval.
\newblock \emph{ICLR}, 2026.

\bibitem[Torne et~al.(2026)Torne, Pertsch, Walke, Vedder, Nair, Ichter, Ren,
  Wang, Tang, Stachowicz, et~al.]{torne2026mem}
Marcel Torne, Karl Pertsch, Homer Walke, Kyle Vedder, Suraj Nair, Brian Ichter,
  Allen~Z. Ren, Haohuan Wang, Jiaming Tang, Kyle Stachowicz, et~al.
\newblock Mem: Multi-scale embodied memory for vision language action models.
\newblock \emph{arXiv preprint arXiv:2603.03596}, 2026.

\bibitem[Wang et~al.(2026)Wang, Qian, Zhang, Xue, Cheng, Wang, Hu, Chen, Shen,
  Huang, and Zhang]{wang2026flashnav}
Shanze Wang, Yiwei Qian, Xinming Zhang, Jun Xue, Siwei Cheng, Xianghui Wang,
  Qingyuan Hu, Yanjun Chen, Xiaoyu Shen, Hailong Huang, and Wei Zhang.
\newblock {FlashNav}: Training deployable robot navigation policies in seconds.
\newblock \emph{arXiv preprint arXiv:2606.15846}, 2026.

\bibitem[Wu et~al.(2025)Wu, Ji, Li, Zhang, He, Xie, Zhang, Bayramli, Ding, and
  Lu]{wu2025dejavu}
Shaokai Wu, Yanbiao Ji, Qiuchang Li, Zhiyi Zhang, Qichen He, Wenyuan Xie,
  Guodong Zhang, Bayram Bayramli, Yue Ding, and Hongtao Lu.
\newblock Dejavu: Towards experience feedback learning for embodied
  intelligence.
\newblock \emph{arXiv preprint arXiv:2510.10181}, 2025.

\bibitem[Xu et~al.(2025)Xu, Wang, Xia, Zhu, Huang, and Xu]{Xu2025VLACache}
Siyu Xu, Yunke Wang, Chenghao Xia, Dihao Zhu, Tao Huang, and Chang Xu.
\newblock Vla-cache: Towards efficient vision-language-action model via
  adaptive token caching in robotic manipulation.
\newblock \emph{NeurIPS}, 2025.

\bibitem[Yang et~al.(2025{\natexlab{a}})Yang, Tan, Wu, Zheng, Peng, Liang, Gu,
  Cai, Ye, Jang, et~al.]{Yang2025a}
Jianwei Yang, Reuben Tan, Qianhui Wu, Ruijie Zheng, Baolin Peng, Yongyuan
  Liang, Yu~Gu, Mu~Cai, Seonghyeon Ye, Joel Jang, et~al.
\newblock Magma: A foundation model for multimodal ai agents.
\newblock \emph{CVPR}, 2025{\natexlab{a}}.

\bibitem[Yang et~al.(2025{\natexlab{b}})Yang, Yu, Wu, Yan, Li, Cheng, Zou,
  Fang, Cheng, Qiu, et~al.]{Yang2025}
Ruihan Yang, Qinxi Yu, Yecheng Wu, Rui Yan, Borui Li, An-Chieh Cheng, Xueyan
  Zou, Yunhao Fang, Xuxin Cheng, Ri-Zhao Qiu, et~al.
\newblock Egovla: Learning vision-language-action models from egocentric human
  videos.
\newblock \emph{arXiv preprint arXiv:2507.12440}, 2025{\natexlab{b}}.

\bibitem[Yang et~al.(2025{\natexlab{c}})Yang, Wang, Wen, Zhongwei, Zou, Zhang,
  Wen, and Zhang]{Yang2025EfficientVLA}
Yantai Yang, Yuhao Wang, Zichen Wen, Luo Zhongwei, Chang Zou, Zhipeng Zhang,
  Chuan Wen, and Linfeng Zhang.
\newblock Efficientvla: Training-free acceleration and compression for
  vision-language-action models.
\newblock \emph{NeurIPS}, 2025{\natexlab{c}}.

\bibitem[Yuan et~al.(2025)Yuan, Liu, Lu, Chen, Jiang, and
  Zhao]{yuan2025depthvla}
Tianyuan Yuan, Yicheng Liu, Chenhao Lu, Zhuoguang Chen, Tao Jiang, and Hang
  Zhao.
\newblock Depthvla: Enhancing vision-language-action models with depth-aware
  spatial reasoning.
\newblock \emph{arXiv preprint arXiv:2510.13375}, 2025.

\bibitem[Zhan et~al.(2026)Zhan, Chen, Zhao, Gao, and
  Yang]{zhan2026directexperienceworldmodeloptimization}
Xiangcheng Zhan, Zirui Chen, Yicheng Zhao, Ziteng Gao, and Shuo Yang.
\newblock Direct experience world-model optimization: Learning the world beyond
  action imitation.
\newblock \emph{arXiv preprint arXiv:2609.37398}, 2026.

\bibitem[Zhang et~al.(2026{\natexlab{a}})Zhang, Li, Dai, Zhu, Zhou, Liu, Wang,
  Tay, Chen, Liu, et~al.]{zhang2025spatial}
Zhengshen Zhang, Hao Li, Yalun Dai, Zhengbang Zhu, Lei Zhou, Chenchen Liu, Dong
  Wang, Francis~EH Tay, Sijin Chen, Ziwei Liu, et~al.
\newblock From spatial to actions: Grounding vision-language-action model in
  spatial foundation priors.
\newblock \emph{ICLR}, 2026{\natexlab{a}}.

\bibitem[Zhang et~al.(2025{\natexlab{a}})Zhang, Dai, Wen, and
  Yang]{zhang2025logosp}
Zihui Zhang, Weisheng Dai, Hongtao Wen, and Bo~Yang.
\newblock {LogoSP}: Local-global grouping of superpoints for unsupervised
  semantic segmentation of {3D} point clouds.
\newblock \emph{CVPR}, 2025{\natexlab{a}}.

\bibitem[Zhang et~al.(2026{\natexlab{b}})Zhang, Dai, Wang, Li, and
  Yang]{zhang2026growsp++}
Zihui Zhang, Weisheng Dai, Bing Wang, Bo~Li, and Bo~Yang.
\newblock {GrowSP++}: Growing superpoints and primitives for unsupervised {3D}
  semantic segmentation.
\newblock \emph{IEEE TPAMI}, 2026{\natexlab{b}}.

\bibitem[Zhang et~al.(2025{\natexlab{b}})Zhang, Xu, Yang, Yue, Lin, Gao, Wang,
  and Zhao]{zhang2025elucidating}
Zongzheng Zhang, Haobo Xu, Zhuo Yang, Chenghao Yue, Zehao Lin, Huan-ang Gao,
  Ziwei Wang, and Hao Zhao.
\newblock Elucidating the design space of torque-aware vision-language-action
  models.
\newblock \emph{CoRL}, 2025{\natexlab{b}}.

\bibitem[Zheng et~al.(2025)Zheng, Liang, Huang, Gao, Daum{\'e}~III, Kolobov,
  Huang, and Yang]{Zheng2024}
Ruijie Zheng, Yongyuan Liang, Shuaiyi Huang, Jianfeng Gao, Hal Daum{\'e}~III,
  Andrey Kolobov, Furong Huang, and Jianwei Yang.
\newblock Tracevla: Visual trace prompting enhances spatial-temporal awareness
  for generalist robotic policies.
\newblock \emph{ICLR}, 2025.

\bibitem[Zhou et~al.(2026{\natexlab{a}})Zhou, Gao, Hong, Liu, Zhang, Dai, Zhen,
  Lyu, Wu, Mao, et~al.]{zhou2026touchanything}
Jianyi Zhou, Ziteng Gao, Feiyang Hong, Zirui Liu, Guannan Zhang, Weisheng Dai,
  Ruichen Zhen, Chuqiao Lyu, Haotian Wu, Yinian Mao, et~al.
\newblock Touchanything: A dataset and framework for bimanual tactile
  estimation from egocentric video.
\newblock \emph{arXiv preprint arXiv:2605.13083}, 2026{\natexlab{a}}.

\bibitem[Zhou et~al.(2026{\natexlab{b}})Zhou, Hong, Li, Zhao, Cen, Liu, Huang,
  Chen, Zhang, Zhu, et~al.]{zhou2026touchworld}
Jianyi Zhou, Feiyang Hong, Yunhao Li, Yicheng Zhao, Yongjue Cen, Zirui Liu,
  Jiakang Huang, Zirui Chen, Ruiyang Zhang, Weizhuo Zhu, et~al.
\newblock Touchworld: A predictive and reactive tactile foundation model for
  dexterous manipulation.
\newblock \emph{arXiv preprint arXiv:2607.07287}, 2026{\natexlab{b}}.

\bibitem[Zhou et~al.(2026{\natexlab{c}})Zhou, Wei, Zhen, Zhao, Xia, Shao, Su,
  and Yang]{zhou2026inject}
Jianyi Zhou, Yujie Wei, Ruichen Zhen, Bo~Zhao, Xiaobo Xia, Rui Shao, Xiu Su,
  and Shuo Yang.
\newblock Inject once survive later: Backdooring vision-language-action models
  to persist through downstream fine-tuning.
\newblock \emph{arXiv preprint arXiv:2602.00500}, 2026{\natexlab{c}}.

\bibitem[Zitkovich et~al.(2023)Zitkovich, Yu, Xu, Xu, Xiao, Xia, Wu, Wohlhart,
  Welker, Wahid, et~al.]{brohan2023rt2}
Brianna Zitkovich, Tianhe Yu, Sichun Xu, Peng Xu, Ted Xiao, Fei Xia, Jialin Wu,
  Paul Wohlhart, Stefan Welker, Ayzaan Wahid, et~al.
\newblock Rt-2: Vision-language-action models transfer web knowledge to robotic
  control.
\newblock \emph{CoRL}, 2023.

\end{thebibliography}
}

\clearpage
\appendix
\section{Additional Implementation Details}

\subsection{Backbone and Fine-tuning Scope}
Our policy follows the VLM+DiT multi-frame post-training setting in the main paper. Fine-tuning freezes the vision encoders and language model, while jointly optimizing the multimodal projector, Learnable History Condensation (LHC), and diffusion action decoder. Historical frame features are detached before LHC; gradients through the current-frame path update the projector. LHC adds trainable capacity and optimization cost, so this setup does not isolate temporal condensation from all effects of added capacity.

At each environment step, the VLM encodes the current observation into a frame-level feature $f_t \in \mathbb{R}^D$. A FIFO buffer stores the most recent $M$ historical features, and the LHC module maps them into $K$ history slots. The decoder is then conditioned on the concatenation $Z_t = [\tilde{F}_t; f_t]$, where $\tilde{F}_t \in \mathbb{R}^{K \times D}$ denotes the condensed history tokens.

\subsection{LHC Parameterization}
Each slot has a base query $q_i^{\mathrm{base}}\in\mathbb{R}^D$ and a slot embedding $e_i^{\mathrm{slot}}\in\mathbb{R}^D$. With $r_i=(i-1)/(K-1)$, its query is $q_i=\mathrm{MLP}([q_i^{\mathrm{base}};r_i])+e_i^{\mathrm{slot}}$. The two-layer MLP uses SiLU; these slot identities do not prescribe which history position a slot attends to. LHC uses four attention heads, with $D=4096$ and $d_h=1024$.

A learned table $E_{\mathrm{time}}\in\mathbb{R}^{M\times D}$, ordered from oldest to newest, enters only the value path with a learnable scale $s$ initialized to zero. Writing $q=[q_1;\ldots;q_K]\in\mathbb R^{K\times D}$ and omitting projection biases,
\begin{equation}
Q=qW_Q,\qquad K_{\mathrm{hist}}=F_tW_K,\qquad V=(F_t+sE_{\mathrm{time}})W_V,
\end{equation}
where $W_Q,W_K,W_V,W_O\in\mathbb R^{D\times D}$. Splitting into $H$ heads gives $Q^{(h)}\in\mathbb R^{K\times d_h}$ and $K_{\mathrm{hist}}^{(h)},V^{(h)}\in\mathbb R^{M\times d_h}$. The head outputs are concatenated and right-multiplied by $W_O$. Keys do not receive the time embedding, so the attention logits have no explicit history-position bias. Table~\ref{tab:appendix_architecture} summarizes the layer dimensions and decoder-conditioning structure.

\subsection{Algorithmic Summary}
Algorithm~\ref{alg:appendix_lhc_inference} summarizes policy prediction before action execution. Let $E_\psi$ denote per-frame VLM encoding, where only the multimodal projector parameters $\psi$ are trainable during fine-tuning. Algorithm~\ref{alg:appendix_lhc_training} optimizes $\psi$, LHC parameters $\Phi$, and decoder parameters $\theta$, with frozen vision encoders and language-model weights. Below, $\mathrm{LHC}_\Phi$ includes the value-only time embedding and the Q/K/V/output projections.

\begin{algorithm}[htbp]
\caption{{\small CondenseVLA Inference with Learnable History Condensation}}
\label{alg:appendix_lhc_inference}
\begin{algorithmic}[1]
\footnotesize
\State \textbf{Input:} $(o_t,x)$, FIFO $\mathcal{B}_{t-1}$, encoder $E_\psi$, LHC $\Phi$, decoder $D_\theta$
\State Retrieve up to $M$ available historical features $F_t$ from $\mathcal{B}_{t-1}$
\State Encode the current step once: $f_t \leftarrow E_\psi(o_t, x)$
\For{$i=1$ to $K$}
\State Form slot query $q_i \leftarrow \mathrm{MLP}([q_i^{\mathrm{base}}; r_i]) + e_i^{\mathrm{slot}}$
\EndFor
\State Stack unprojected queries as $q=[q_1;\ldots;q_K]$
\If{$\mathcal{B}_{t-1}$ is empty}
\State $\tilde{F}_t\leftarrow\mathrm{Repeat}(f_t,K)$
\Else
\State $\tilde{F}_t\leftarrow\mathrm{LHC}_\Phi(q,F_t)$ using available-history time embeddings
\EndIf
\State Form decoder conditioning tokens: $Z_t \leftarrow [\tilde{F}_t; f_t]$
\State Predict the action chunk with the diffusion decoder: $\hat{a}_{t:t+T_a-1} \leftarrow D_\theta(Z_t)$
\State Update the FIFO buffer: $\mathcal{B}_t \leftarrow \mathrm{PushFIFO}(\mathcal{B}_{t-1}, f_t)$
\State \textbf{Return:} $\hat{a}_{t:t+T_a-1}, \mathcal{B}_t$
\end{algorithmic}
\end{algorithm}

\begin{algorithm}[htbp]
\caption{{\small Fine-tuning CondenseVLA}}
\label{alg:appendix_lhc_training}
\begin{algorithmic}[1]
\footnotesize
\State \textbf{Input:} observation-window mini-batch, encoder $E_\psi$, LHC $\Phi$, decoder $D_\theta$
\For{each mini-batch}
\State Encode each frame with $E_\psi$; split $f_t$ and history $F_t\leftarrow\mathrm{sg}(F_t)$
\For{$i=1$ to $K$}
\State Form slot query $q_i \leftarrow \mathrm{MLP}([q_i^{\mathrm{base}}; r_i]) + e_i^{\mathrm{slot}}$
\EndFor
\State Stack unprojected queries as $q=[q_1;\ldots;q_K]$
\State Compute slots and per-head maps: $(\tilde{F}_t,\{A^{(h)}\}_{h=1}^{H}) \leftarrow \mathrm{LHC}_\Phi(q,F_t)$
\State Form decoder tokens: $Z_t \leftarrow [\tilde{F}_t; f_t]$
\State Sample diffusion timestep $\tau$ and Gaussian noise $\epsilon$
\State Construct the noised action chunk $a^\epsilon_{t:t+T_a-1}$
\State Predict noise: $\hat{\epsilon} \leftarrow D_\theta(a^\epsilon_{t:t+T_a-1}, Z_t; \tau)$
\State Compute diffusion loss $L_{\mathrm{diff}}\leftarrow\mathrm{mean}((\epsilon-\hat{\epsilon})^2)$
\State Average attention over heads, then the local mini-batch; normalize rows to obtain $\bar A$
\State $L_{\mathrm{div}}\leftarrow K^{-2}\|\bar A\bar A^\top-I\|_F^2$
\State Compute $L_{\mathrm{slot}}$ using the normalized pair weights in Eq.~\ref{eq:slot_weighted}
\State Form the total objective $L \leftarrow L_{\mathrm{diff}} + \lambda_{\mathrm{div}}L_{\mathrm{div}} + \lambda_{\mathrm{slot}}L_{\mathrm{slot}}$
\State Update $\psi,\Phi,\theta$; keep vision-encoder and language-model weights frozen
\EndFor
\end{algorithmic}
\end{algorithm}

\subsection{Training Details}
\label{sec:appendix_loss_reductions}
For Simpler-WidowX, LIBERO, and real-world experiments, the learning rate is fixed to $2\times 10^{-5}$. The auxiliary weights are $\lambda_{\mathrm{div}}=0.06$ and $\lambda_{\mathrm{slot}}=0.001$. The multimodal projector, LHC, and diffusion decoder are optimized under
\begin{equation}
L = L_{\mathrm{diff}} + \lambda_{\mathrm{div}} L_{\mathrm{div}} + \lambda_{\mathrm{slot}} L_{\mathrm{slot}}.
\end{equation}
The implementation uses AdamW with zero weight decay, a constant learning rate without warm-up, and gradient-norm clipping at 1.0. The training entry point defaults to seed 42; this training seed is distinct from the repeated evaluation seeds. Diffusion training uses 100 timesteps with the \texttt{squaredcos\_cap\_v2} noise schedule and mean squared noise-prediction error over batch, action positions, and action dimensions. The Simpler and LIBERO training launchers use four independently noised diffusion samples per conditioning example. Evaluation uses 10-step DDIM with $\eta=0$ and Gaussian initial noise.

\paragraph{Loss reductions.}
The MSE notation in the main paper includes entry-wise averaging. For a predicted noise chunk $\hat\epsilon$, action horizon $T_a$, and action dimension $d_a$, the expanded losses are
\begin{equation}
\begin{aligned}
L_{\mathrm{diff}}&=\mathbb E_{\epsilon,\tau}\left[\frac{\|\epsilon-\hat\epsilon\|_2^2}{T_a d_a}\right],\\
L_{\mathrm{div}}&=\frac{\|\bar A\bar A^\top-I\|_F^2}{K^2}.
\end{aligned}
\end{equation}
The main-paper slot-pair mean uses the normalized reduction in Eq.~\ref{eq:slot_weighted}; its denominator protection returns zero when no pair is active. These are the same reductions used with the reported auxiliary weights.

For Simpler and LIBERO, the single-arm action representation has seven components: three end-effector translation components, three rotation components, and a gripper command. Dataset-specific 1st/99th percentile statistics normalize the masked continuous dimensions to $[-1,1]$; inference clips predictions to this range and inverts the normalization, with gripper conversion in the benchmark wrapper. The Simpler and LIBERO implementations predict 16 and 8 actions per chunk, respectively, including the current action. LIBERO fine-tuning uses global batch 64 and per-device batch 16 on four processes. Tables~\ref{tab:appendix_hparams} and~\ref{tab:appendix_hparams_libero} summarize the benchmark-specific training settings.

The benchmark-specific data and history settings are summarized below:
\begin{itemize}
\item \textbf{Simpler-WidowX}~\cite{li2024evaluating}: models are trained on the \texttt{bridge\_rt\_1}+Fractal data mixture, with $M=20$ historical features and $K=6$ history slots.
\item \textbf{LIBERO}~\cite{liu2023libero}: models are trained on LIBERO data, with $M=10$ historical features and $K=3$ history slots. Our LIBERO model uses external and wrist RGB inputs and does not use state input.
\item \textbf{Real-world Galaxea R1 Lite}: our policy is trained with real-robot data using the LIBERO-style optimization flow, with $M=20$ and $K=6$, while the input modality follows the Simpler setting without state input or wrist-view input.
\end{itemize}

Tables~\ref{tab:appendix_hparams} and~\ref{tab:appendix_hparams_libero} report training hyperparameters for Simpler-WidowX, real-world tasks, and the separately fine-tuned LIBERO suites. Table~\ref{tab:appendix_architecture} summarizes the architecture and reported parameter counts.
\begin{table}[htbp]
\centering
\caption{Run-level training hyperparameters for Simpler-WidowX and real-world experiments.}
\label{tab:appendix_hparams}
\begin{tabular*}{\linewidth}{@{\extracolsep{\fill}}lcc}
\toprule
Setting & Simpler-WidowX & Real-world \\
\midrule
Learning rate & $2\times 10^{-5}$ & $2\times 10^{-5}$ \\
History length $M$ & $20$ & $20$ \\
Number of slots $K$ & $6$ & $6$ \\
Batch size & $64$ & $64$ \\
Training epochs & $4$ & $17$ \\
Diffusion steps at inference & $10$ & $10$ \\
\bottomrule
\end{tabular*}
\end{table}

\begin{table}[htbp]
\centering
\caption{Suite-wise LIBERO fine-tuning hyperparameters. Each suite is fine-tuned separately from the same multi-frame post-training initialization.}
\label{tab:appendix_hparams_libero}
\begin{tabular*}{\linewidth}{@{\extracolsep{\fill}}lcccc}
\toprule
Setting & Spatial & Object & Goal & Long \\
\midrule
Learning rate & $2\times 10^{-5}$ & $2\times 10^{-5}$ & $2\times 10^{-5}$ & $2\times 10^{-5}$ \\
History length $M$ & $10$ & $10$ & $10$ & $10$ \\
Number of slots $K$ & $3$ & $3$ & $3$ & $3$ \\
Global batch size & $64$ & $64$ & $64$ & $64$ \\
Training epochs & $13$ & $21$ & $18$ & $25$ \\
Diffusion steps at inference & $10$ & $10$ & $10$ & $10$ \\
\bottomrule
\end{tabular*}
\end{table}

\subsection{Architecture and Parameter Count}
\label{sec:appendix_protocol_details}
This subsection collects implementation details that are useful for reproducibility but too verbose for the main paper. \Cref{tab:appendix_architecture} summarizes the architecture and parameter counts. Unless otherwise noted, the parameter counts are reported for the Simpler-WidowX setting with $M=20$ and $K=6$.

\begin{table}[htbp]
\centering
\caption{Module summary for $M=20,K=6$. Vision encoders and the language model are frozen; the separately listed MM projector, LHC, and decoder are trainable. Counts are reported totals, with LHC verified analytically from its layer definitions. Here $B$ is batch size, $F$ encoded frames, $P$ visual tokens, $T$ text tokens, $A$ action horizon, $D=4096$, and $d=768$.}
\label{tab:appendix_architecture}
\scriptsize
\begingroup
\setlength{\tabcolsep}{2pt}
\begin{tabularx}{\linewidth}{@{}p{0.22\linewidth}p{0.19\linewidth}p{0.12\linewidth}X X@{}}
\toprule
Module (Parameters) & Submodule & Layer Num & Input Shape & Output Shape \\
\midrule
VLM backbone (7.47B) & SigLIP vision encoder & $1+27+1$ & $(B \cdot F, 3, H, W)$ & $(B \cdot F, P, 1152)$ \\
 & DINOv2 vision encoder & $1+24+1$ & $(B \cdot F, 3, H, W)$ & $(B \cdot F, P, 1024)$ \\
 & LLaMA2 language backbone & $1+32+0$ & $(B \cdot F, P+T, D)$ & $(B, F, D)$ \\
\midrule
MM projector (71.4M) & No-align fused-GELU MLP & $1$ & $(B \cdot F, P, 2176)$ & $(B \cdot F, P, D)$ \\
\midrule
LHC (100.8M) & Base queries + slot embeddings & $1$ & two $(K,D)$ tables & $(K,D)$ \\
 & Query MLP (SiLU) & $2$ linear & $(K,D+1)$ & $(K,D)$ \\
 & Value time embedding + scale & $1$ & $(M,D)$, scalar & $(B,M,D)$ \\
 & Q/K/V/O projections & $1$ & $(B,K,D)$, $(B,M,D)$ & $(B,K,D)$, $(B,M,D)$ \\
 & Cross-attention & $1$ & $(B,K,D)$, $(B,M,D)$ & $(B,K,D)$ \\
\midrule
History path (0) & FIFO feature buffer & -- & $(B,F,D)$ & $(B,M,D)$ \\
 & History-current fusion & -- & $(B,K,D)$, $(B,1,D)$ & $(B,K+1,D)$ \\
\midrule
Diffusion action decoder (134.7M) & Action embedding & $1$ & $(B,A,7)$ & $(B,A,d)$ \\
 & Timestep embedding & $1$ & $(B,256)$ & $(B,1,d)$ \\
 & Conditioning embedder & $1$ & $(B,K+1,D)$ & $(B,K+6,d)$ \\
 & Self-attn. + MLP & $12$ & $(B,A+1,d)$ & $(B,A+1,d)$ \\
 & Cross-attn. & $12$ & $(B,A+1,d)$, $(B,K+6,d)$ & $(B,A+1,d)$ \\
 & Final layer & $1$ & $(B,A+1,d)$ & $(B,A,7)$ \\
\midrule
\MethodName\ total (7.78B) & Full policy & -- & RGB observation, text, cached history & $(B,A,7)$ \\
\bottomrule
\end{tabularx}
\endgroup
\end{table}

\section{Benchmark Setup}

\subsection{Simpler-WidowX: Evaluation, Uncertainty, and Provenance}
\label{sec:appendix_simpler}
We use the standard prepackaged SimplerEnv WidowX visual-matching setting, not SimplerEnv-OR. The four tasks are Spoon on Towel, Carrot on Plate, Stack Cube, and Eggplant in Basket. Evaluation uses episode IDs 0--23, \texttt{control\_freq=5}, \texttt{sim\_freq=500}, \texttt{max\_episode\_steps=120}, and the standard task-success criterion. These tasks primarily probe manipulation robustness under the simulated visual and dynamics gap, rather than general-purpose memory.

For each of the local CronusVLA and \MethodName\ checkpoints, we repeat evaluation with three evaluation seeds on the same 24 episode IDs, yielding 72 rollouts per task and 288 across the four tasks. These are fixed-checkpoint evaluations, not three independently trained models or 72 distinct scene IDs. Table~\ref{tab:appendix_simpler_uncertainty} reports the mean and sample standard deviation (SD) across evaluation seeds. For Avg., the four-task average is computed within each seed before the across-seed SD; task SDs are not averaged. The original 24-rollout experiments are reported separately as detailed grasp/success results and ablations. No uncertainty from the final-checkpoint reevaluation is assigned to other ablation variants.

\begin{table}[htbp]
\centering
\caption{Simpler fixed-checkpoint evaluation: mean $\pm$ evaluation-seed SD (\%), and pooled Wilson 95\% intervals. Each task uses $3\times24$ rollouts; Avg. pools 288 rollouts for its interval.}
\label{tab:appendix_simpler_uncertainty}
\small
\begin{tabular*}{\linewidth}{@{\extracolsep{\fill}}lcccc}
\toprule
& \multicolumn{2}{c}{CronusVLA} & \multicolumn{2}{c}{\MethodName} \\
\cmidrule(lr){2-3}\cmidrule(lr){4-5}
Task & Mean $\pm$ SD & 95\% CI & Mean $\pm$ SD & 95\% CI \\
\midrule
Spoon on Towel & $66.7\pm0.0$ & [55.2, 76.5] & $80.6\pm6.4$ & [70.0, 88.0] \\
Carrot on Plate & $51.4\pm8.7$ & [40.1, 62.6] & $66.7\pm0.0$ & [55.2, 76.5] \\
Stack Cube & $22.2\pm6.4$ & [14.2, 33.1] & $33.3\pm8.3$ & [23.5, 44.8] \\
Eggplant in Basket & $94.4\pm6.4$ & [86.6, 97.8] & $98.6\pm2.4$ & [92.5, 99.8] \\
Avg. & $58.7\pm1.6$ & [52.9, 64.2] & $69.8\pm1.0$ & [64.3, 74.8] \\
\bottomrule
\end{tabular*}
\end{table}

\paragraph{Detailed grasp and success results.}
Table~\ref{tab:appendix_simpler_detailed} reports the original grasp/success breakdown. Its local results are from the original single 24-rollout evaluation, rather than the three-seed reevaluation used in the main paper and Table~\ref{tab:appendix_simpler_uncertainty}.

\begin{table*}[htbp]
\centering
\caption{Original Simpler-WidowX detailed results (\%): grasp and final success. The local CronusVLA and CondenseVLA rows use the original 24-rollout-per-task evaluation; Avg. averages final success.}
\label{tab:appendix_simpler_detailed}
\scriptsize
\begin{tabular*}{\textwidth}{@{\extracolsep{\fill}}lccccccccc}
\toprule
& \multicolumn{2}{c}{\shortstack{Spoon\\on Towel}} & \multicolumn{2}{c}{\shortstack{Carrot\\on Plate}} & \multicolumn{2}{c}{\shortstack{Stack\\Cube}} & \multicolumn{2}{c}{\shortstack{Eggplant\\in Basket}} & \\
\cmidrule(lr){2-3} \cmidrule(lr){4-5} \cmidrule(lr){6-7} \cmidrule(lr){8-9}
Method & Grasp & Success & Grasp & Success & Grasp & Success & Grasp & Success & Avg. \\
\midrule
RT-1-X~\cite{ONeill2024} & 16.7 & 0.0 & 20.8 & 4.2 & 8.3 & 0.0 & 0.0 & 0.0 & 1.1 \\
RoboVLMs~\cite{Li2024c} & 70.8 & 50.0 & 45.8 & 37.5 & 70.8 & 0.0 & 91.7 & 83.3 & 42.7 \\
Magma~\cite{Yang2025a} & 62.5 & 37.5 & 41.7 & 29.2 & 79.2 & 20.8 & 95.8 & 91.7 & 44.8 \\
SpatialVLA~\cite{Qu2025} & 25.0 & 20.8 & 41.7 & 37.5 & 79.2 & \textbf{41.7} & 91.7 & 83.3 & 45.8 \\
HAMLET~\cite{Koo2025HAMLET} & \textbf{91.7} & 75.0 & \textbf{83.3} & 62.5 & 75.0 & 16.7 & \textbf{100.0} & \textbf{100.0} & 63.5 \\
\midrule
OpenVLA~\cite{Kim2024} & 12.5 & 8.3 & 29.2 & 4.2 & 16.7 & 0.0 & 20.8 & 0.0 & 3.1 \\
CronusVLA~\cite{Li2025} & 79.2 & 66.7 & 70.8 & 41.7 & 58.3 & 29.2 & \textbf{100.0} & 95.8 & 58.4 \\
\textbf{CondenseVLA} & 87.5 & \textbf{79.2} & \textbf{83.3} & \textbf{66.7} & \textbf{83.3} & \textbf{41.7} & \textbf{100.0} & 95.8 & \textbf{70.8} \\
\bottomrule
\end{tabular*}
\end{table*}

The breakdown separates object acquisition from full task completion. The gap is particularly pronounced on Stack Cube, where successful pickup must be followed by precise alignment and contact-sensitive placement. CondenseVLA achieves strong grasp and final-success rates on Spoon on Towel and Carrot on Plate. On Eggplant in Basket, the strongest methods already reach saturated grasp rates, while final success still depends on reliable transport and placement.

\paragraph{Interval definition and scope.}
For $s$ successes in $n$ rollouts, let $\hat p=s/n$ and $z=1.96$. The Wilson interval used here and for the real-world results is
\begin{equation}
\label{eq:wilson_interval}
\frac{\hat p+z^2/(2n)\;\pm\;z\sqrt{\hat p(1-\hat p)/n+z^2/(4n^2)}}{1+z^2/n},
\end{equation}
with endpoints multiplied by 100. The pooled interval treats trials with a binomial model, conditional on the checkpoint and evaluation setting. It does not capture training randomness, and repeated scene IDs or cross-task heterogeneity are not fully represented by the pooled model. We report these intervals descriptively and do not infer paired method differences or task-wise significance from interval overlap.

\paragraph{Baseline sources.}
The main-paper Simpler results table combines local training/evaluation, local reproductions, and external reports. \MethodName\ is trained and evaluated locally; OpenVLA-OFT and CronusVLA are local reproductions. Only the CronusVLA and \MethodName\ rows are replaced by the three-seed reevaluation means. TraceVLA, Octo, CogACT, Magma, RT-1-X, SpatialVLA, GR00T-N1.5, and OpenVLA are taken from the CronusVLA comparison. UniVLA uses its corrected GitHub release; $\pi_0$ and ContextVLA use the ContextVLA report; MemoryVLA and HAMLET use their respective papers. In particular, the OpenVLA row is an external reference, not a local current-frame-only ablation. Table grouping does not imply identical training or evaluation protocols.

External methods retain their reported data and input protocols: for example, $\pi_0$ and GR00T-N1.5 additionally use robot state, while UniVLA uses robot and human-video training data. We do not attach the local evaluation intervals to these external results. The original CronusVLA setting uses six dense frame features; the dense 10/20/30-frame variants used in our scaling study are retrained for four epochs with matched initialization, training data, optimization, and evaluation. Those variants are distinct from the original baseline and from the matched $M=20$ RMBench comparison.

\subsection{LIBERO}
\label{sec:appendix_libero}
% Camera-ready V03: three evaluation repeats confirmed by the author.
% Camera-ready V03: original +/- values confirmed by the author as evaluation-seed SD.
LIBERO~\cite{liu2023libero} contains four suites, Spatial, Object, Goal, and Long, with ten tasks in each suite. We fine-tune from multi-frame post-training weights separately for each suite. Our model uses external and wrist RGB inputs without state input. Local LIBERO evaluations repeat each fixed trained checkpoint with three evaluation seeds, not three independently trained models. Each seed runs 50 rollouts per task, giving 500 episodes per suite and seed, 1,500 per suite over three repeats, and 6,000 across the four suites per local method. Suite means average tasks and evaluation repeats; the overall mean weights the four suites equally. The main-paper table reports means only; Table~\ref{tab:appendix_libero_protocol_results} reports mean $\pm$ standard deviation (SD) across the three evaluation seeds for \MethodName. The Avg. uncertainty describes the SD of the per-repeat overall success rates. External rows retain their source protocols and are not assigned this local evaluation budget.

Because recent LIBERO results are reported under different input protocols, \cref{tab:appendix_libero_protocol_results} summarizes the input setting together with the suite-level numbers used in the main paper. CronusVLA and \MethodName\ use external and wrist RGB inputs without state input, while some reported baselines, such as \(\pi_0\) and OpenVLA-OFT, use both wrist-view and robot-state inputs.

\begin{table}[htbp]
\centering
\caption{LIBERO success rates (\%) under different input protocols. \MethodName\ reports mean $\pm$ SD across three evaluation seeds of fixed trained checkpoints; other rows report means. The $\pi_{0.5}$ average retains the source precision.}
\label{tab:appendix_libero_protocol_results}
\scriptsize
\begingroup
\setlength{\tabcolsep}{2pt}
\begin{tabular*}{\linewidth}{@{\extracolsep{\fill}}llccccc}
\toprule
Method & Input setting & Spatial & Object & Goal & Long & Avg. \\
\midrule
Diffusion Policy~\cite{Chi2024DiffusionPolicy} & no state, no wrist & 78.3 & 92.5 & 68.3 & 50.5 & 72.4 \\
Dita~\cite{Hou2025Dita} & no state, no wrist & 84.2 & 96.3 & 85.4 & 63.8 & 82.4 \\
OpenVLA~\cite{Kim2024} & no state, no wrist & 84.7 & 88.4 & 79.2 & 53.7 & 76.5 \\
TraceVLA~\cite{Zheng2024} & no state, no wrist & 84.6 & 85.2 & 75.1 & 54.1 & 74.8 \\
SpatialVLA~\cite{Qu2025} & no state, no wrist & 88.2 & 89.9 & 78.6 & 55.5 & 78.1 \\
OpenVLA-OFT~\cite{Kim2025} & wrist + state & 97.6 & 98.4 & 97.9 & 94.5 & 97.1 \\
$\pi_0$~\cite{Black2024} & wrist + state & 96.8 & 98.8 & 95.8 & 85.2 & 94.2 \\
$\pi_{0.5}$ (openpi, reported) & wrist + state & \textbf{98.8} & 98.2 & \textbf{98.0} & 92.4 & 96.85 \\
CronusVLA~\cite{Li2025} & wrist, no state & 97.1 & 98.1 & 96.0 & 94.0 & 96.3 \\
\textbf{\MethodName} & wrist, no state & $97.8\pm0.8$ & $\mathbf{99.1\pm0.5}$ & $97.4\pm0.9$ & $\mathbf{94.6\pm1.1}$ & $\mathbf{97.2\pm0.8}$ \\
\bottomrule
\end{tabular*}
\endgroup
\end{table}

The $\pi_{0.5}$ reference is the \texttt{pi05\_libero} checkpoint at 30k steps from the \href{https://github.com/Physical-Intelligence/openpi/blob/main/examples/libero/README.md}{official openpi LIBERO release}, with external and wrist RGB observations and 8-D proprioception. Both use external and wrist RGB observations, but our policy does not use proprioceptive state and the training protocols differ; it is therefore a contextual reference. Our main same-protocol comparison is CronusVLA (96.3\%) versus \MethodName\ (97.2\%). LIBERO-Long involves multi-step execution, but the near-saturated scores do not by themselves isolate a memory mechanism.

\subsection{RMBench: Protocol, Task Results, and Scope}
\label{sec:appendix_rmbench}
We use RMBench's nine memory-dependent tasks~\cite{chen2026rmbench} with 50 training demonstrations and 100 evaluation rollouts per task. We follow RMBench's \texttt{demo\_clean} environment configuration with the Aloha-AgileX dual-arm robot in RoboTwin~2.0/SAPIEN. Our policies use head and wrist RGB observations without proprioceptive state input and output end-effector commands for both arms. Dense CronusVLA and \MethodName\ share the same initialization, data, training steps, seed, action horizon, and evaluation protocol. Both read the same 20 consecutive historical features from a rolling FIFO, with the current observation supplied separately. \MethodName\ maps the historical features to six slots. The $M=20,K=6$ configuration is transferred from Simpler without task-specific retuning on RMBench; no separate held-out hyperparameter split was used to select it. The current-frame-only OpenVLA result is a reference for the absence of explicit history, while our primary controlled comparison is between dense and condensed history conditioning.

\begin{table}[htbp]
\centering
\caption{Full RMBench success rates (\%). Each task has 100 rollouts, so each task-level percentage also gives the number of successful rollouts. The category averages weight tasks equally; Overall weights all nine tasks equally.}
\label{tab:appendix_rmbench_tasks}
\small
\begin{tabular*}{\linewidth}{@{\extracolsep{\fill}}llrrr}
\toprule
Category & Task & OpenVLA & CronusVLA & \MethodName \\
\midrule
$M(1)$ & Observe and Pick Up & 0.0 & 5.0 & 12.0 \\
 & Rearrange Blocks & 5.0 & 23.0 & 39.0 \\
 & Put Back Block & 9.0 & 35.0 & 46.0 \\
 & Swap Blocks & 10.0 & 21.0 & 35.0 \\
 & Swap T & 3.0 & 7.0 & 12.0 \\
 & Average & 5.4 & 18.2 & 28.8 \\
\midrule
$M(n)$ & Battery Try & 10.0 & 12.0 & 15.0 \\
 & Blocks Ranking Try & 5.0 & 9.0 & 13.0 \\
 & Cover Blocks & 0.0 & 17.0 & 22.0 \\
 & Press Button & 0.0 & 0.0 & 0.0 \\
 & Average & 3.8 & 9.5 & 12.5 \\
\midrule
All tasks & Overall & 4.7 & 14.3 & 21.6 \\
\bottomrule
\end{tabular*}
\end{table}

The dense and condensed policies succeed in 129/900 and 194/900 rollouts, respectively, giving an overall improvement of $100(194-129)/900=7.22$ pp before rounding. The largest task-level gains occur on Rearrange Blocks (+16 pp), Swap Blocks (+14 pp), and Put Back Block (+11 pp). The smaller improvement on $M(n)$ and the zero success on Press Button indicate that bounded visual conditioning does not solve persistent multi-observation accumulation.

For context, \href{https://arxiv.org/html/2603.01229v2}{RMBench v2, Table 1}~\cite{chen2026rmbench} reports Mem-0 success of 52.8\% on $M(1)$ and 28.5\% on $M(n)$. These are externally reported results from a system with a different backbone, persistent memory, and high-level planning. They are not a matched comparison with our local policies and are excluded from the controlled table. The difference highlights the scope of \MethodName: low-level conditioning over recent cached features, with no persistent semantic memory or hierarchical planner.

\subsection{Real-world Galaxea R1 Lite}
Real-world evaluation is performed on a Galaxea R1 Lite robot with a third-person RGB camera mounted near the top center of the robot body. \MethodName\ uses the LIBERO-style optimization flow, with $M=20$ and $K=6$, while retaining the Simpler-style input modality without state input or wrist-view input. We consider two simple tasks, Put Apple on Plate and Open then Close Drawer, and three long-horizon tasks, Pick and Place Twice, Swap Cubes, and Put Objects in Drawer. For each task, 50 real-world trajectories are collected for training, and success rate is computed over 50 evaluation trials per method.

\begin{figure}[htbp]
\centering
\includegraphics[width=0.82\linewidth,height=0.42\textheight,keepaspectratio]{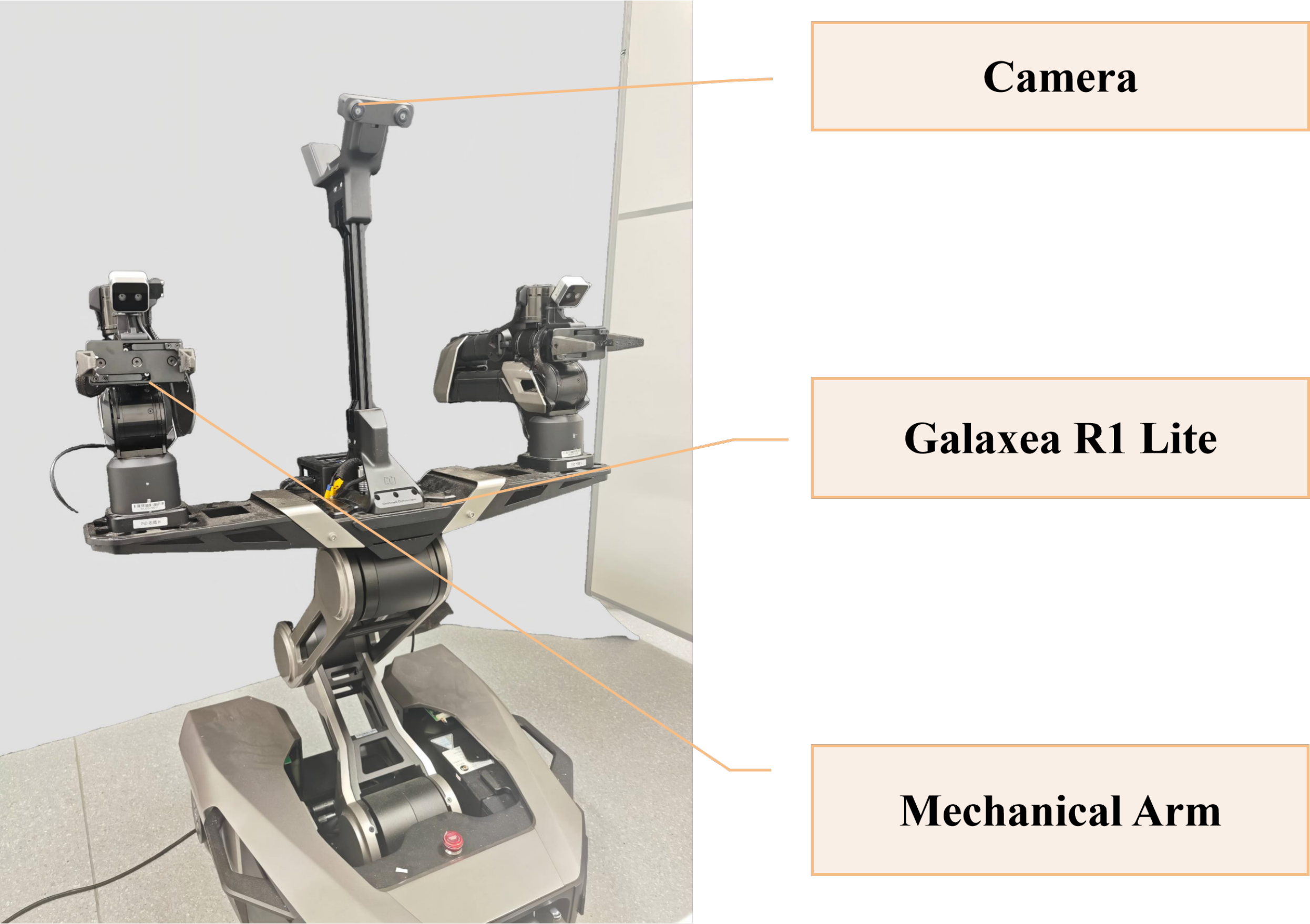}
\caption{Real-world deployment setup on the Galaxea R1 Lite platform, showing the camera, robot body, and manipulator used in our experiments.}
\label{fig:appendix_real_setup}
\end{figure}

This setup jointly evaluates low-level manipulation accuracy and short-term temporal context under multi-step execution. The simple tasks primarily probe immediate control quality, while the long-horizon tasks require the policy to track recently completed subtasks, object order, and task-stage transitions. In practice, the long-horizon tasks are substantially more sensitive to stage-tracking errors, contact uncertainty, and execution drift. A visual overview of the platform, camera layout, and evaluated tasks is provided in \cref{fig:appendix_real_setup}.

The real-world tasks differ not only in manipulation difficulty but also in their temporal structure. \textbf{Put Apple on Plate} is a short-horizon task with a simple structure of grasping and placement, so success mainly depends on immediate perception and execution accuracy. \textbf{Open then Close Drawer} introduces a two-stage structure, requiring the policy to distinguish whether the current subgoal is opening or closing and to re-establish handle interaction after the first stage. \textbf{Pick and Place Twice} requires placing the object onto the other plate and then returning it to the original plate, so the policy must track the current task stage across recent observations. \textbf{Swap Cubes} requires exchanging the positions of two objects, which places additional demand on tracking object identity and operation order across multiple stages. \textbf{Put Objects in Drawer} requires placing both the cube and apple inside the drawer and then closing it. The intermediate metric records the first object placement; final success requires both insertions and drawer closure, so insertion alone is insufficient. Detailed stage-wise outcomes and failure modes are discussed in \cref{sec:appendix_realworld_discussion}.

\section{Additional Ablations}

\subsection{Regularization Variants}
The main paper studies the overall contribution of LHC components. Here, we examine the design of the slot-redundancy regularizer and compare several variants for selecting or weighting slot pairs. Let $\tilde{f}_q \in \mathbb{R}^{D}$ denote the output embedding of slot $q$, and let $\hat{A}_{q,:} \in \mathbb{R}^{M}$ denote the head-averaged and row-normalized slot-to-history attention of slot $q$, where $q \in \{1,\dots,K\}$ and $M$ is the history length. We define the cosine similarity between two slot embeddings as
\begin{equation}
c_{ij} = \cos(\tilde{f}_i,\tilde{f}_j),
\end{equation}
and the cosine similarity between their attention rows as
\begin{equation}
a_{ij} = \cos(\hat{A}_{i,:}, \hat{A}_{j,:}).
\end{equation}

For a per-device mini-batch of size $B$, we write the slot decorrelation loss in a unified form:
\begin{equation}
\label{eq:slot_weighted}
L_{\mathrm{slot}} =
\frac{1}{Z}
\sum_{b=1}^{B}\sum_{1 \le i < j \le K}
w_{b,ij}\, c_{b,ij}^{2},
\qquad
Z = \max\!\left(\sum_{b=1}^{B}\sum_{i<j} w_{b,ij},\varepsilon\right),
\end{equation}
where $w_{b,ij}\geq0$ selects or weights pairs, and $\varepsilon=10^{-6}$. Both numerator and denominator sum over samples and pairs; this is not an unweighted average of separately normalized sample losses when active weights differ across samples. Cosine similarities are computed in FP32 with norm protection and clamped to $[-1,1]$ before squaring. The denominator clamp makes the expression well-defined for any active pair set. The reported middle-only configuration uses no additional all-pairs term. Below we omit $b$ when describing a single sample's weights.

For attention diversity, $\hat A_b=H^{-1}\sum_h A_b^{(h)}$ is averaged across the local mini-batch, then each row is normalized with a denominator floor of $10^{-8}$ to obtain $\bar A$. The loss is $K^{-2}\|\bar A\bar A^\top-I\|_F^2$, computed after these averages, rather than averaging sample-wise Gram losses. Its off-diagonal entries discourage overlap, while its diagonal entries encourage concentrated attention rows. The outer weights $\lambda_{\mathrm{div}}$ and $\lambda_{\mathrm{slot}}$ multiply these reduced losses.

\paragraph{\textbf{All-pairs.}}
The all-pairs variant penalizes all off-diagonal slot pairs uniformly:
\begin{equation}
w_{ij}^{\text{all}} = 1, \qquad 1 \le i < j \le K.
\end{equation}
This is the most direct form of slot decorrelation and imposes global redundancy reduction across all slots.

\paragraph{\textbf{Middle-only.}}
The middle-only variant excludes the first and last slot indices and penalizes only pairs among the remaining slots:
\begin{equation}
w_{ij}^{\text{mid}} =
\mathbf{1}[\,2 \le i < j \le K-1\,].
\end{equation}
This mask leaves the first and last slot indices unconstrained by this term. It does not impose a temporal-endpoint role on either slot.

\paragraph{\textbf{Dynamic middle masking.}}
We first estimate the mean temporal position of each slot using its normalized attention row:
\begin{equation}
\mu_q = \sum_{m=1}^{M} m \, \hat{A}_{q,m}.
\end{equation}
Here, $\mu_q$ is the attention-weighted history index. For each sample, let $q_{\min}=\arg\min_q\mu_q$ and $q_{\max}=\arg\max_q\mu_q$, using the first index to break ties. We remove these indices and set $\mathcal{S}_{\mathrm{mid}}=\{1,\ldots,K\}\setminus\{q_{\min},q_{\max}\}$. The decorrelation weights are
\begin{equation}
w_{ij}^{\text{dyn-mid}} =
\mathbf{1}[\,i \in \mathcal{S}_{\mathrm{mid}}\,]\,
\mathbf{1}[\,j \in \mathcal{S}_{\mathrm{mid}}\,].
\end{equation}
Unlike fixed middle-only masking, membership depends on each sample's attention centroids. When the extrema select distinct slots, both masks retain four slots and six pairs at $K=6$. If all centroids are equal, both extrema select the same index and the dynamic mask removes only that index; Eq.~\ref{eq:slot_weighted} normalizes by the actual active-pair count.

\paragraph{\textbf{Gate-by-attention.}}
Instead of selecting pairs by slot index or mean temporal position, this variant weights slot pairs according to attention-row similarity:
\begin{equation}
g_{ij} = \mathrm{ReLU}(a_{ij} - \tau),
\end{equation}
where $\tau$ is a similarity threshold. In our experiments, we set $\tau = 0.6$. We stop gradients through the gate and define
\begin{equation}
w_{ij}^{\text{gate}} = \mathrm{sg}(g_{ij}),
\end{equation}
where $\mathrm{sg}(\cdot)$ denotes stop-gradient (detach). Thus, pairs with highly similar attention patterns are penalized more strongly, while dissimilar pairs contribute little or no decorrelation penalty.

Thus, \textbf{all-pairs} constrains every pair, \textbf{middle-only} selects a fixed set by slot index, \textbf{dynamic middle masking} selects by sample-dependent centroids, and \textbf{gate-by-attention} weights pairs by attention similarity. These variants compare pair selection or weighting within the same LHC architecture.

Unless otherwise noted, the results in this ablation are reported on Simpler-WidowX under the same setting as the main-text regularization study, with history length $M=20$, slot number $K=6$, $\lambda_{\mathrm{div}}=0.06$, and $\lambda_{\mathrm{slot}}=0.001$. All variants retain the attention-diversity term $L_{\mathrm{div}}$; the only difference lies in how the slot-pair weights $w_{ij}$ are defined in $L_{\mathrm{slot}}$. Average success rate is computed over the four Simpler tasks, with 24 evaluation rollouts per task.

\begin{table}[htbp]
\centering
\caption{Comparison of slot-redundancy reduction variants on Simpler-WidowX. $\Delta=\mathrm{SR}_{\mathrm{variant}}-\mathrm{SR}_{\mathrm{middle-only}}$ is a signed difference in percentage points (pp).}
\label{tab:appendix_reg_ablation}
\begin{tabular*}{\linewidth}{@{\extracolsep{\fill}}lccc}
\toprule
Variant & Pair rule & Avg. success (\%) & $\Delta$ (pp) \\
\midrule
All-pairs & all off-diagonal pairs & 68.8 & -2.0 \\
\textbf{Middle-only} & fixed interior indices & \textbf{70.8} & 0.0 \\
Dynamic middle masking & sample-adaptive middle slots & 52.1 & -18.7 \\
Gate-by-attention & attention-similarity gating & 64.6 & -6.2 \\
\bottomrule
\end{tabular*}
\end{table}

The fixed middle-only variant has the highest reported success in \cref{tab:appendix_reg_ablation}. Dynamic middle masking performs worse despite normally retaining the same number of slots and pairs. The outcome favors fixed membership in this setting, but success rates alone do not establish training instability or explain why particular overlaps help or hurt. These results do not establish fixed temporal roles for boundary slot indices.

\subsection{Slot Importance Under Slot Zeroing}
To further probe whether the learned slots play interchangeable roles, we perform a slot-zero ablation on the best \textit{middle-only} checkpoint. At inference time, we zero out one slot at a time while keeping the remaining slots unchanged, and then reevaluate the policy on Simpler-WidowX. This experiment reveals how much the final policy depends on each individual slot after training has converged.

\begin{figure}[htbp]
\centering
\includegraphics[width=0.82\linewidth]{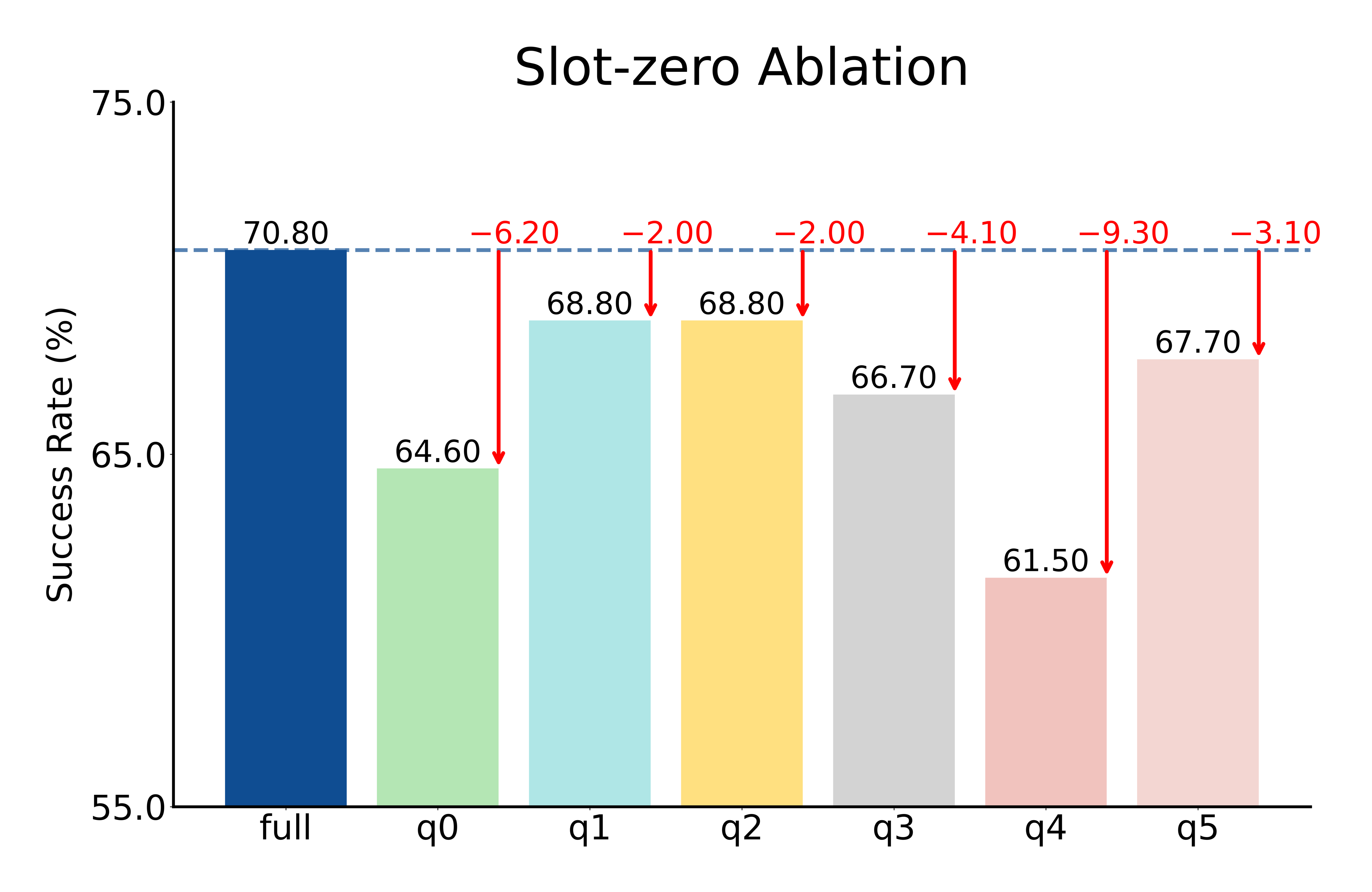}
\caption{Slot-zero ablation on Simpler-WidowX. Each bar shows the success rate after zeroing one slot at inference time; the dashed line indicates the full-model result. The uneven drop across slots suggests that the learned slots are not fully interchangeable, while the limited degradation for some slots still indicates partial redundancy.}
\label{fig:appendix_slotzero}
\end{figure}

The results in \cref{fig:appendix_slotzero} show unequal sensitivity to removing individual slots. Zeroing plotted slot $q_4$ (mathematical index $i=5$) causes the largest drop, while removing some other slots has a smaller effect. This is a dependence test for the evaluated checkpoint, not evidence that a slot has a fixed temporal role or that its importance transfers across tasks and training runs. It also does not by itself establish why the middle-only regularizer performs better.

\subsection{Sensitivity to History Length and Condensation Strategy}
The history-length analysis in the main paper already shows that performance peaks at a moderate temporal window. In the supplementary material, we therefore focus on two complementary comparisons: (i) how naive dense-history extension behaves in the CronusVLA~\cite{Li2025} baseline, and (ii) how learned condensation compares against non-learnable condensed-history baselines under the same history and slot budget. The corresponding comparisons are reported in \cref{tab:appendix_history_perf_cost,tab:appendix_handcrafted_history}.

\begin{table}[htbp]
\centering
\caption{History-length success rates on Simpler-WidowX under the original 24-rollout-per-task ablation protocol. Runtime is reported separately in Table~\ref{tab:appendix_latency_breakdown}.}
\label{tab:appendix_history_perf_cost}
\begin{tabular*}{\linewidth}{@{\extracolsep{\fill}}lcc}
\toprule
Method & History setting & Avg. success (\%) \\
\midrule
CronusVLA~\cite{Li2025} & original 6-frame setting & 58.4 \\
CronusVLA~\cite{Li2025} & $M=10$ & 57.3 \\
CronusVLA~\cite{Li2025} & $M=20$ & 53.1 \\
CronusVLA~\cite{Li2025} & $M=30$ & 47.9 \\
\MethodName & $M=10$, $K=6$ & 63.5 \\
\MethodName & $M=20$, $K=6$ & \textbf{70.8} \\
\MethodName & $M=30$, $K=6$ & 60.4 \\
\bottomrule
\end{tabular*}
\end{table}

\begin{table}[htbp]
\centering
\caption{History-selection comparisons at matched history and output budgets. Static-Index and Static-Weight replace LHC at inference without retraining. Gap is variant SR minus learned SR, in percentage points (pp).}
\label{tab:appendix_handcrafted_history}
\begin{tabular*}{\linewidth}{@{\extracolsep{\fill}}lccc}
\toprule
Method & Construction & Avg. success (\%) & Gap to learned (pp) \\
\midrule
Static-Index Selection & fixed endpoint gather & 55.2 & -15.6 \\
Static-Weight Selection & fixed endpoint kernel & 54.2 & -16.6 \\
Strided History Sampling & fixed interval gather & 59.4 & -11.4 \\
\textbf{CondenseVLA (learned)} & input-adaptive attention & \textbf{70.8} & 0.0 \\
\bottomrule
\end{tabular*}
\end{table}

Within the evaluated policy family, Table~\ref{tab:appendix_history_perf_cost} shows that simply extending dense conditioning does not improve success. The condensed policy performs best at $M=20$ for fixed $K=6$, while the full $M\times K$ grid shows that increasing $K$ helps at $M=30$. These observations support joint selection of history and slot budgets; they do not establish a universal redundancy principle or isolate the reason for lower success at a longer window.

For \cref{tab:appendix_handcrafted_history}, let $F_t=\{f_{t-M},\ldots,f_{t-1}\}$ and $K=6$. Static-Index and Static-Weight are inference-time replacements in a trained policy, without retraining the decoder. Their endpoint templates follow the coarse endpoint preference observed in learned attention. Both bypass the complete LHC, including its value-time embeddings and learned projections, and operate directly on cached frame features. \textbf{Static-Index Selection} gathers the six positions
\begin{equation}
[1,\;1,\;M,\;M,\;M,\;M].
\end{equation}
This baseline therefore forces two slots to copy the oldest feature and four slots to copy the most recent feature, without any input-dependent adaptation. In other words, the template is intended to preserve the same coarse ``early anchor + strong recent emphasis'' pattern observed in the learned model, while removing any state-dependent adjustment of where the slots attend.

\textbf{Static-Weight Selection} keeps the same endpoint bias but replaces hard gathering with fixed local kernels. Concretely, for each slot $q$, we define a hand-crafted weight distribution $p_q(m)$ over history index $m \in \{1,\dots,M\}$ and compute the slot feature as
\begin{equation}
\tilde{f}_q = \sum_{m=1}^{M} p_q(m)\, f_{t-M-1+m},
\qquad
\sum_{m=1}^{M} p_q(m)=1.
\end{equation}
For ``old'' slots, the kernel is centered near the oldest position; for ``last'' slots, it is centered near the most recent position. We use a triangular kernel
\begin{equation}
p_q(m) \propto \max\!\left(0,\, w_q+1-|m-c_q|\right),
\end{equation}
followed by row normalization, with $c_q=1,w_q=1$ for the two old slots and $c_q=M,w_q=2$ for the four recent slots. Thus, the old endpoint uses relative weights $(2,1)$ and the recent endpoint uses $(1,2,3)$ from older to newer positions, truncated to the available history. These are the implementation's integer-width parameters.

\textbf{Strided History Sampling} is a simple temporal downsampling baseline. It keeps the same history budget $K=6$ as CondenseVLA but selects six fixed positions from the $M=20$ history window, namely
\begin{equation}
[0,\;4,\;8,\;11,\;15,\;19],
\end{equation}
using zero-based indexing over the cached history. This baseline tests whether the gain can be explained by uniformly reducing the number of history features rather than by input-adaptive condensation.

The tested alternatives in \cref{tab:appendix_handcrafted_history} do not match learned condensation; the reported strided result trails it by 11.4 pp. The endpoint interventions test replacing the learned representation at inference, including a change to the decoder's conditioning distribution. They are not independently retrained, capacity-matched baselines and cannot isolate input-adaptive attention as the sole source of the gain.

\subsection{Slot-Regularization Strength}
Table~\ref{tab:appendix_sensitivity_best} reports best-checkpoint success rates for two alternative slot-regularization weights. The main setting uses $\lambda_{\mathrm{slot}}=0.001$.

\begin{table}[htbp]
\centering
\caption{Reported best-checkpoint success rates at two alternative slot-regularization weights on Simpler-WidowX.}
\label{tab:appendix_sensitivity_best}
\begin{tabular*}{\linewidth}{@{\extracolsep{\fill}}lcc}
\toprule
Group & Setting & Best SR (\%) \\
\midrule
Slot regularization & slot-regularization weight = 0.005 & 64.6 \\
Slot regularization & slot-regularization weight = 0.01 & 58.3 \\
\bottomrule
\end{tabular*}
\end{table}

\subsection{Policy Latency and Component Costs}
\label{sec:appendix_latency}
We profile one NVIDIA A100-SXM4-80GB at batch size 1 with 10-step DDIM, using 40 Simpler episodes (10 per task), 10 warm-up calls, and steady-state histories. All rows in Table~\ref{tab:appendix_latency_breakdown} use the same device, timers, synchronization, and profiling protocol. The component sum is VLM + LHC + decoder. Measured total directly times the complete policy forward, including conditioning, fusion, synchronization, and other regions; it is not inferred from the component sum or the reciprocal of a reported frequency. Policy-forward latency is distinct from the physical closed-loop control period.

\begin{table}[htbp]
\centering
\caption{Policy latency in ms per step under one profiling protocol. Boldface marks lower aggregate values within matched $M$ pairs. The original $M=6$ dense setting is an unmatched reference.}
\label{tab:appendix_latency_breakdown}
\small
\setlength{\tabcolsep}{3pt}
\begin{tabular*}{\linewidth}{@{\extracolsep{\fill}}llrrrrr}
\toprule
Method & History & VLM & LHC & Decoder & Sum & Total \\
\midrule
Dense CronusVLA & $M=6$ & 108.37 & 0.00 & 106.00 & 214.37 & 214.89 \\
\midrule
Dense CronusVLA & $M=10$ & 108.11 & 0.00 & 110.09 & 218.20 & \textbf{219.07} \\
\MethodName & $M=10,K=6$ & 108.42 & 2.26 & 106.44 & \textbf{217.12} & 219.62 \\
\midrule
Dense CronusVLA & $M=20$ & 108.28 & 0.00 & 120.32 & 228.60 & 229.52 \\
\MethodName & $M=20,K=6$ & 108.15 & 4.52 & 106.85 & \textbf{219.52} & \textbf{220.88} \\
\midrule
Dense CronusVLA & $M=30$ & 107.94 & 0.00 & 130.54 & 238.48 & 239.97 \\
\MethodName & $M=30,K=6$ & 108.19 & 6.79 & 106.14 & \textbf{221.12} & \textbf{222.14} \\
\bottomrule
\end{tabular*}
\end{table}

At matched $M=10$, \MethodName\ has a 0.5\% lower component sum but a 0.25\% higher measured total. At $M=20$ and $M=30$, measured totals are 3.8\% and 7.4\% lower, respectively. VLM cost remains near 108 ms because only the current observation is encoded. With $K=6$, decoder cost stays near 106--107 ms, but LHC cost grows from 2.26 to 6.79 ms. The measured advantage is bounded decoder conditioning and flatter scaling with history length, not a large or universal end-to-end speedup.

\section{Additional Qualitative and Mechanistic Analysis}

\subsection{Occlusion-based Visualization}
\begin{figure}[htbp]
  \centering
  \setlength{\tabcolsep}{2pt}
  \begin{tabular}{ccc}
    \includegraphics[width=0.22\linewidth]{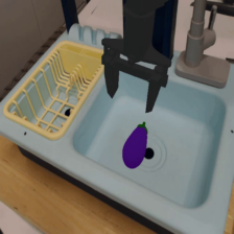} &
    \includegraphics[width=0.22\linewidth]{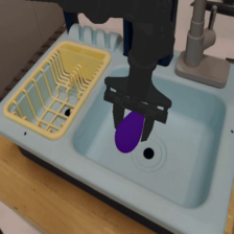} &
    \includegraphics[width=0.22\linewidth]{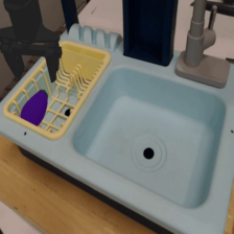} \\
    \includegraphics[width=0.22\linewidth]{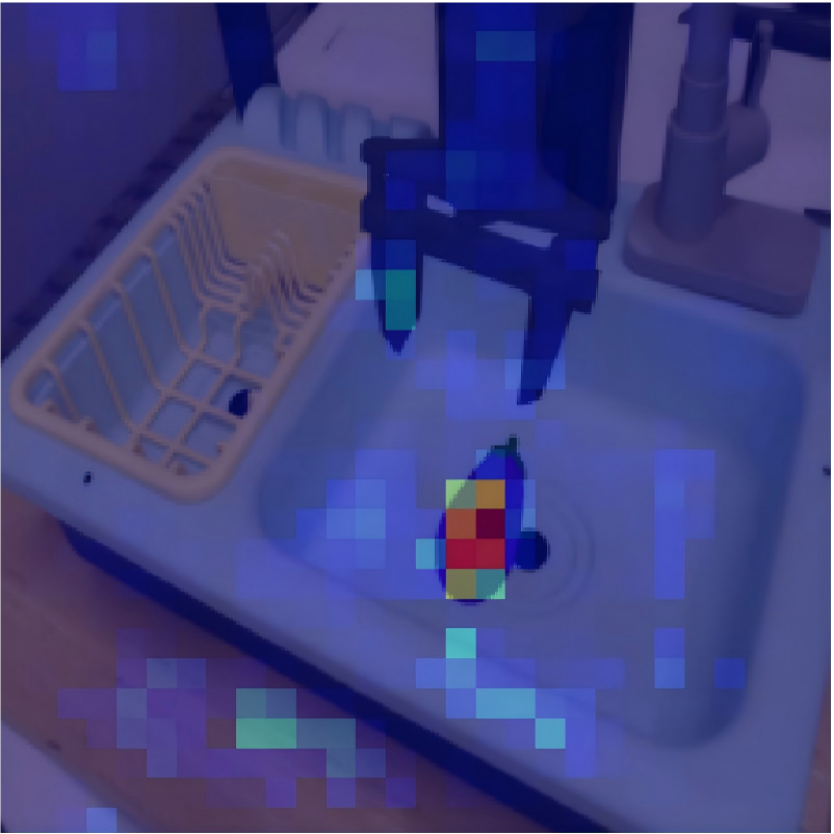} &
    \includegraphics[width=0.22\linewidth]{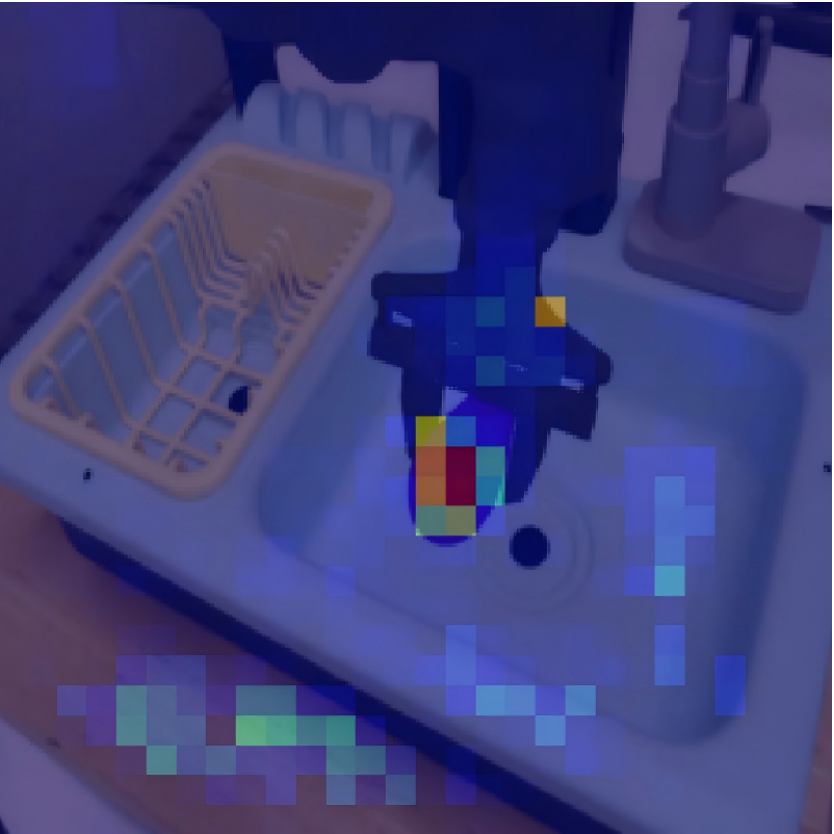} &
    \includegraphics[width=0.22\linewidth]{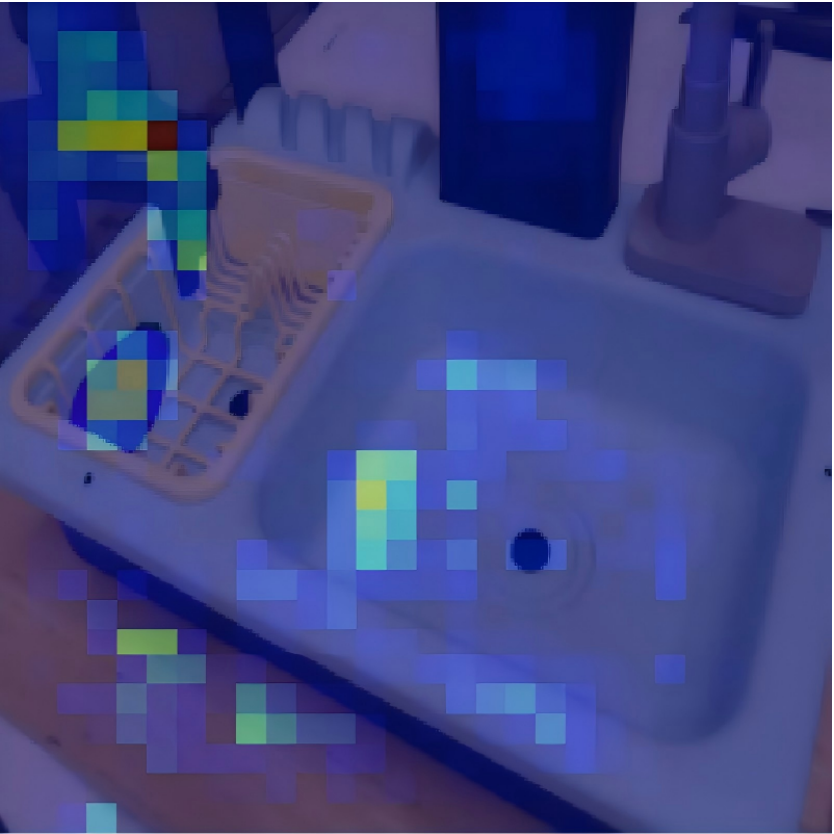} \\
    (a) Early stage & (b) Contact stage & (c) Placement stage
  \end{tabular}
  \caption{Occlusion overlays on SimplerEnv (Eggplant-in-Basket). Highlighted regions show larger cosine-distance changes in the cognition feature under local occlusion. Each heatmap is normalized separately, so colors do not compare absolute sensitivity across frames.}
  \label{fig:simpler_overlay_vis}
\end{figure}

Representative occlusion overlays are shown in \cref{fig:simpler_overlay_vis}. Holding the instruction fixed, we slide a patch over the RGB frame and recompute the VLM cognition feature for each occluded image. For original feature $f$ and occluded feature $f^{\mathrm{occ}}$, the response is $1-\cos(f,f^{\mathrm{occ}})$. Each response is assigned to its patch region; overlapping contributions are averaged per pixel and clamped below at zero. The map is min--max normalized within each image and overlaid with opacity 0.45. This measures sensitivity of the per-frame representation, not a change in predicted actions.

The highlighted regions align with the manipulated object, target receptacle, and nearby contact areas, showing that the current representation is sensitive to task-relevant spatial evidence. Across the early, contact, and placement stages, the overlays provide an intuitive view of the visual regions involved in the ongoing interaction. This spatial diagnostic complements the slot-level temporal analysis below: the overlays probe where the current-frame representation is sensitive, while the slot summaries show when information is retrieved from recent history.

\subsection{Task-level Slot Attention Summaries}
\begin{figure}[htbp]
  \centering
  \includegraphics[width=0.98\linewidth]{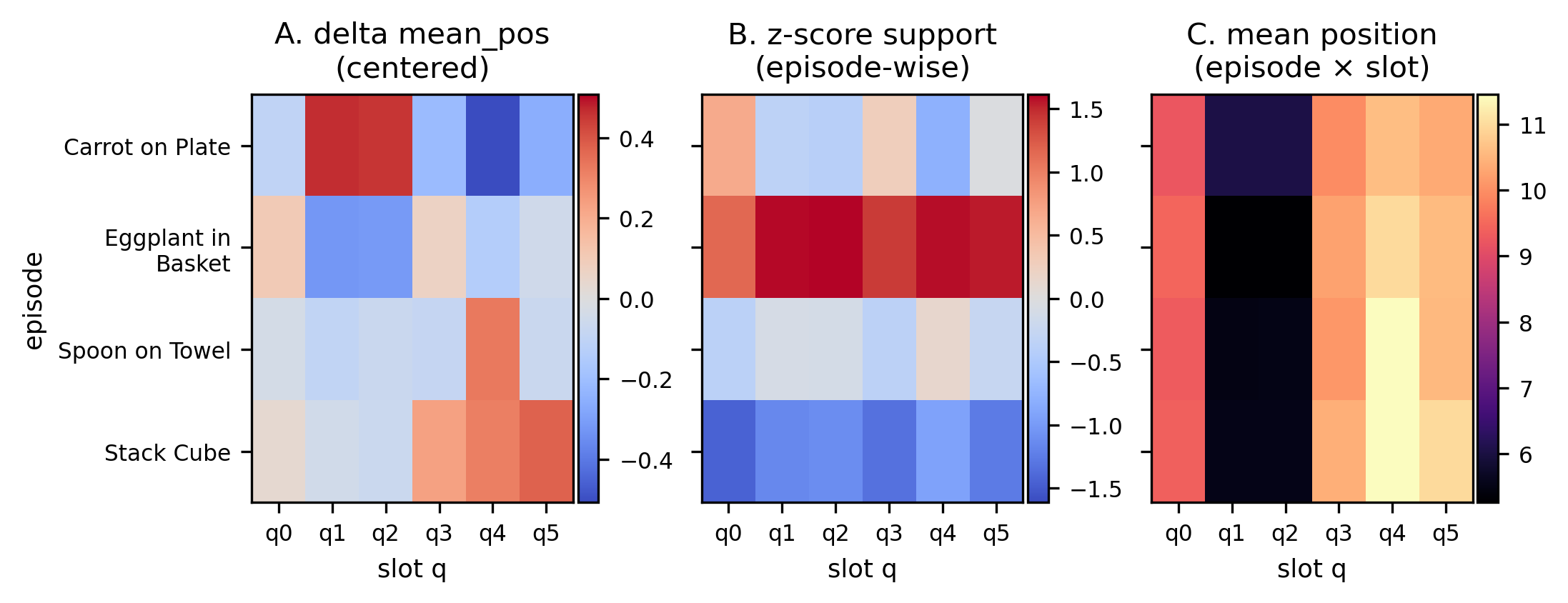}
  \caption{Task-grouped attention summaries (``episode'' in the image refers to a task group). A: per-slot centroid shifts relative to the across-task mean. B: mean effective support standardized across tasks, separately for each slot. C: mean temporal centroids. The views show task-dependent redistribution of temporal position and attention coverage.}
  \label{fig:lhc_triptych_mechanism}
\end{figure}

\begin{figure}[htbp]
\centering
\includegraphics[width=0.98\linewidth]{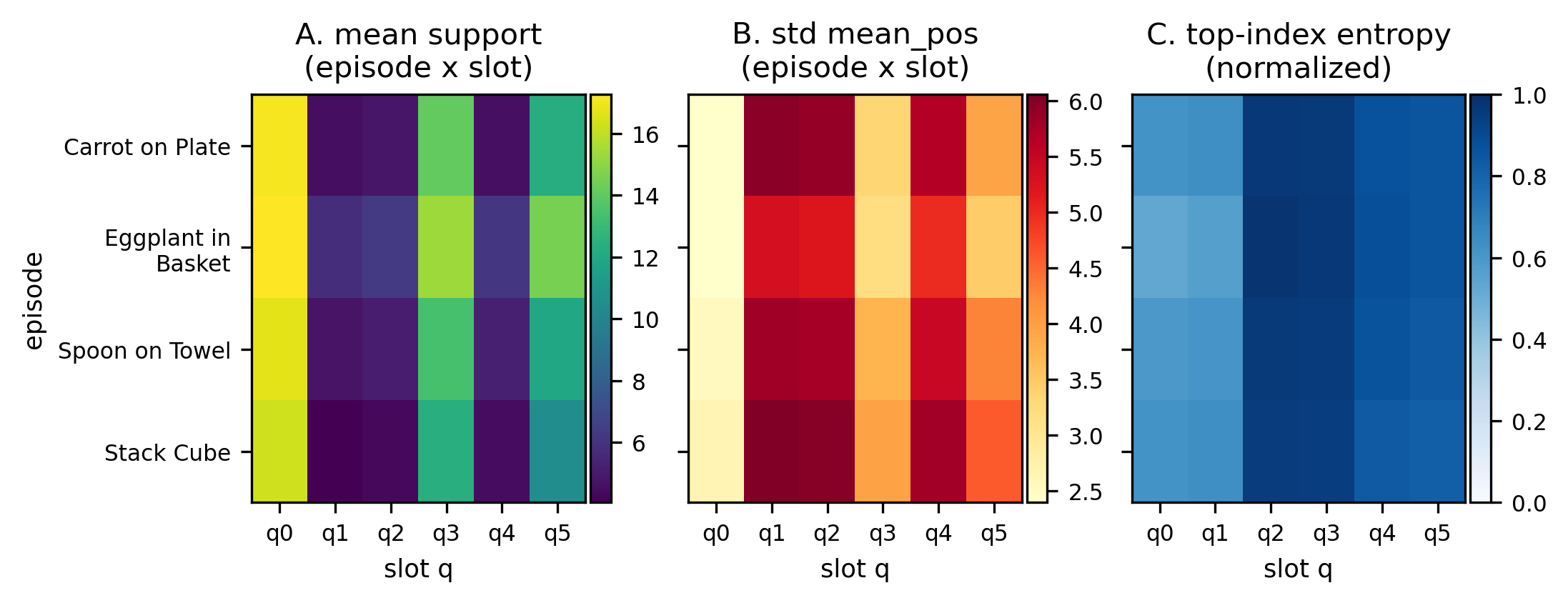}
\caption{Task-grouped slot statistics on Simpler-WidowX (image rows labeled ``episode'' are task groups). A: mean per-step effective support. B: standard deviation of per-step temporal centroids within each task group. C: normalized entropy of top-attended history indices. Together, the panels illustrate structured but flexible temporal retrieval across tasks.}
\label{fig:appendix_stability_summary}
\end{figure}

We summarize logged decision-step attention by task description. Each row in Figures~\ref{fig:lhc_triptych_mechanism} and~\ref{fig:appendix_stability_summary} pools records for one task; ``episode'' in the plots refers to a task group.

For normalized, head-averaged attention $\hat A_{n,q,j}$ at logged step $n$, slot $q$, and zero-based history index $j$, we compute
\begin{equation}
\mu_{n,q}=\sum_{j=0}^{m_n-1}j\hat A_{n,q,j},\qquad
s_{n,q}=\left(\sum_{j=0}^{m_n-1}\hat A_{n,q,j}^{2}\right)^{-1},
\end{equation}
where $m_n$ is the available history length. For task group $g$, $\bar\mu_{g,q}$ and $\bar s_{g,q}$ average these per-step quantities. Thus, mean support is the mean of per-step inverse concentrations, not the inverse concentration of an averaged attention row. In Figure~\ref{fig:lhc_triptych_mechanism}, panel A subtracts the across-task mean separately for each slot, $\Delta\mu_{g,q}=\bar\mu_{g,q}-G^{-1}\sum_{g'}\bar\mu_{g',q}$, where $G=4$. Panel B standardizes $\bar s_{g,q}$ across the four tasks using the population standard deviation (floored at $10^{-9}$); panel C shows $\bar\mu_{g,q}$ directly.

Figure~\ref{fig:appendix_stability_summary} adds mean per-step support, the population standard deviation of per-step centroids within each task group, and top-index entropy. For the latter, each slot's top-attended indices form an empirical distribution within the task group; its entropy is divided by $\log R$, where $R$ is one plus the largest top index in the logged dataset, with a minimum of two. Slots $q_0,q_3,q_5$ have broader support than $q_1,q_2,q_4$ in this run, and $q_0$ has relatively low centroid variation. Lower top-index entropy indicates a more concentrated preference for particular history positions, while higher entropy indicates more variable retrieval. Together, these views show complementary broad-coverage and focused-retrieval patterns, with task-dependent adjustments in temporal position and coverage.

\section{Real-world Discussion and Failure Cases}
\label{sec:appendix_realworld_discussion}

Although CondenseVLA improves over the baselines in real-world evaluation, the absolute success rates remain lower than those in simulation. This gap is expected because real-robot deployment introduces multiple coupled sources of uncertainty, including perception noise, imperfect calibration, unmodeled contact dynamics, execution drift, and scene variation beyond the training distribution~\cite{qian2026eida,wang2026flashnav}. Long-horizon tasks are especially sensitive to these effects because a small early error can propagate into a later stage that is less frequently observed during training.

\begin{table}[htbp]
  \caption{Success rate (\%) over 50 real-world trials per task and method. All intermediate and final metrics use the same 50-trial denominator; Avg. averages final success only. Put Objects in Drawer requires closure after both insertions.}
  \label{tab:appendix_realworld_results}
  \centering
  \scriptsize
  \setlength{\tabcolsep}{1.5pt}
  \begin{tabular*}{\linewidth}{@{\extracolsep{\fill}}lcccc|ccccccc}
    \toprule
    Method & \multicolumn{2}{c}{\shortstack{Put Apple\\on Plate}} & \multicolumn{2}{c|}{\shortstack{Open then\\Close Drawer}} & \multicolumn{2}{c}{\shortstack{Pick and\\Place Twice}} & \multicolumn{2}{c}{\shortstack{Swap\\Cubes}} & \multicolumn{2}{c}{\shortstack{Put Objects\\in Drawer}} & \shortstack{Avg.\\Success} \\
    \cmidrule(lr){2-3} \cmidrule(lr){4-5} \cmidrule(lr){6-7} \cmidrule(lr){8-9} \cmidrule(lr){10-11}
    & Grasp & Success & Open & Success & PnP Once & Success & First Cube & Success & Obj.~1 Placed & Success & \\
    \midrule
    OpenVLA-OFT~\cite{Kim2025} & 52.0 & 48.0 & 56.0 & 44.0 & 48.0 & 16.0 & 48.0 & 8.0 & 12.0 & 0.0 & 23.2 \\
    $\pi_0$~\cite{Black2024} & \textbf{72.0} & \textbf{72.0} & \textbf{68.0} & \textbf{60.0} & \textbf{56.0} & 24.0 & 64.0 & 12.0 & 24.0 & 16.0 & 36.8 \\
    CronusVLA~\cite{Li2025} & 68.0 & 66.0 & 52.0 & 48.0 & 46.0 & 40.0 & 58.0 & 50.0 & 28.0 & 22.0 & 45.2 \\
    \textbf{\MethodName} & \textbf{72.0} & 70.0 & 56.0 & 54.0 & 52.0 & \textbf{50.0} & \textbf{66.0} & \textbf{62.0} & \textbf{32.0} & \textbf{28.0} & \textbf{52.8} \\
    \bottomrule
  \end{tabular*}
\end{table}

The stage-wise metrics in \cref{tab:appendix_realworld_results} show non-uniform improvements of 4--12 pp over CronusVLA. Final success on the three multi-stage tasks improves from 40\% to 50\% on Pick and Place Twice, 50\% to 62\% on Swap Cubes, and 22\% to 28\% on Put Objects in Drawer. Their mean changes from 37.3\% to 46.7\% (+9.3 pp, computed from counts before rounding). In the drawer task, the intermediate metric improves from 28\% to 32\% and records the first object insertion; final success additionally requires the second insertion and drawer closure. These outcomes are consistent with a benefit from recent temporal context, but do not isolate task-stage inference from low-level control accuracy.

\begin{table}[htbp]
\centering
\caption{Real-world final success over 50 trials per task. Counts, rates (\%), and Wilson 95\% intervals are reported for CronusVLA and \MethodName. Aggregate pools 250 trials across the five tasks.}
\label{tab:appendix_realworld_uncertainty}
\small
\begin{tabular*}{\linewidth}{@{\extracolsep{\fill}}lcccc}
\toprule
& \multicolumn{2}{c}{CronusVLA} & \multicolumn{2}{c}{\MethodName} \\
\cmidrule(lr){2-3}\cmidrule(lr){4-5}
Task & Successes (\%) & 95\% CI & Successes (\%) & 95\% CI \\
\midrule
Put Apple on Plate & 33/50 (66.0) & [52.2, 77.6] & 35/50 (70.0) & [56.2, 80.9] \\
Open then Close Drawer & 24/50 (48.0) & [34.8, 61.5] & 27/50 (54.0) & [40.4, 67.0] \\
Pick and Place Twice & 20/50 (40.0) & [27.6, 53.8] & 25/50 (50.0) & [36.6, 63.4] \\
Swap Cubes & 25/50 (50.0) & [36.6, 63.4] & 31/50 (62.0) & [48.2, 74.1] \\
Put Objects in Drawer & 11/50 (22.0) & [12.8, 35.2] & 14/50 (28.0) & [17.5, 41.7] \\
Aggregate & 113/250 (45.2) & [39.1, 51.4] & 132/250 (52.8) & [46.6, 58.9] \\
\bottomrule
\end{tabular*}
\end{table}

Intervals use Eq.~\ref{eq:wilson_interval}, conditional on the evaluated policies and deployment setting. The aggregate is a descriptive pooled binomial interval and does not fully account for task heterogeneity or potential dependence among physical trials. These intervals do not establish paired differences or significance on every task. All methods use 50 evaluation trials per task, and the training budget remains 50 demonstrations per task.

We observe three representative failure patterns. First, the policy can occasionally commit to an incorrect task stage after an ambiguous grasp or partial placement, indicating that the retained history is still insufficient in some corner cases. Second, the policy may correctly infer the task stage but fail at precise low-level execution, especially in drawer interaction or cluttered placement settings. Third, in multi-object tasks, the policy may preserve the correct subgoal but recover too slowly from a perturbed intermediate state, leading to cascading failure in later steps.

These observations suggest several directions for future work: adaptive history length rather than a fixed window, tighter coupling between condensed temporal memory and object-centric grounding, stronger recovery-oriented training under perturbed execution, and broader cross-embodiment validation beyond the current backbone and robot platform. We summarize the dominant real-world failure modes for each task below.

\noindent\textbf{Put Apple on Plate.}
The dominant failure mode is target-object grasp failure: the gripper does not secure the apple reliably before the placement stage, typically due to inaccurate object localization or unstable grasp execution.

\par\noindent\textbf{Open then Close Drawer.}
The main failure mode is handle-grasp failure. During opening, the policy may fail to localize or grasp the handle correctly, which prevents the drawer from being opened. During closing, the same failure can cause the robot to push the drawer front directly instead of re-establishing a stable handle-based interaction.

\par\noindent\textbf{Pick and Place Twice.}
The typical failure mode is wrong subgoal selection: the policy places the object on the wrong target plate or commits to an incorrect intermediate stage, indicating imperfect tracking of task order and stage progression.

\par\noindent\textbf{Swap Cubes.}
This task most often fails through wrong subgoal selection as well, where the policy grasps the wrong cube and then returns it, suggesting that object identity and swap order are not fully preserved under execution noise.

\par\noindent\textbf{Put Objects in Drawer.}
Two failure modes are common. First, the object may collide with the drawer edge during insertion and drop instead of settling inside, reflecting limited contact robustness and placement precision. Second, the policy may exhibit the same handle-grasp failure mode observed in Open then Close Drawer when interacting with the drawer itself.

\begin{figure}[htbp]
\centering
\includegraphics[width=0.98\linewidth]{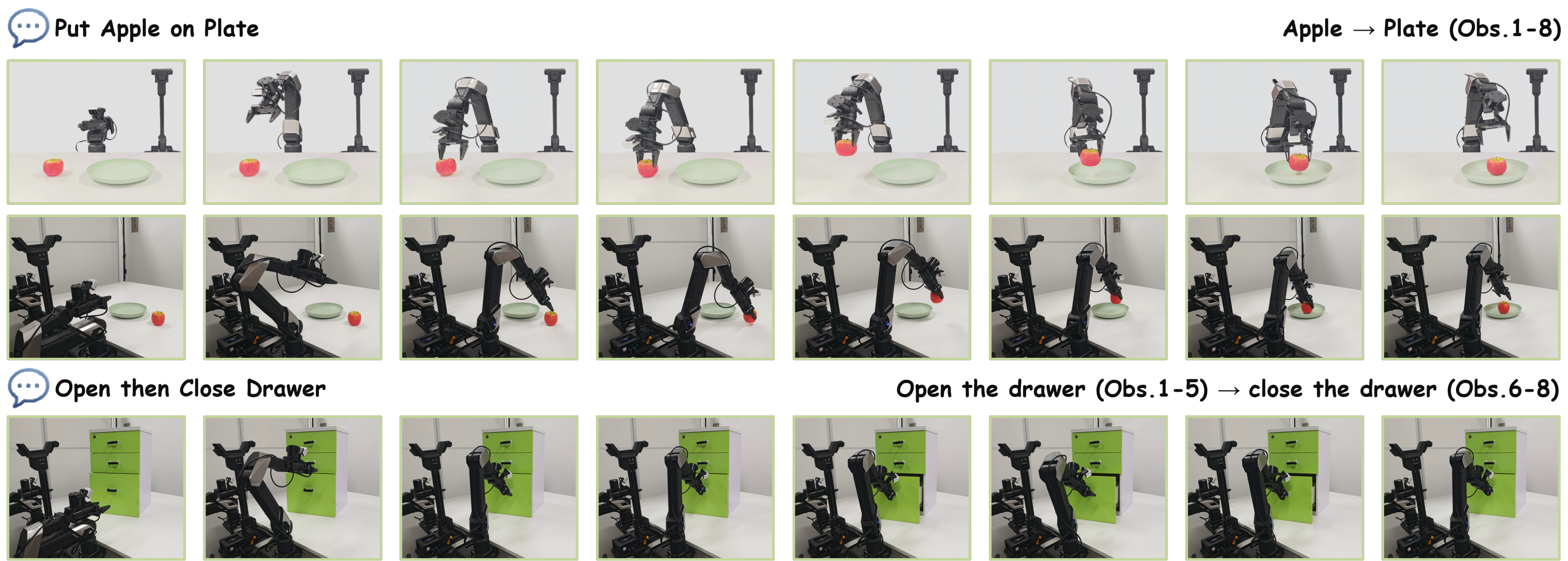}
\par\vspace{0.25em}
{\small (a) Simple tasks}
\par\vspace{0.8em}
\includegraphics[width=0.98\linewidth]{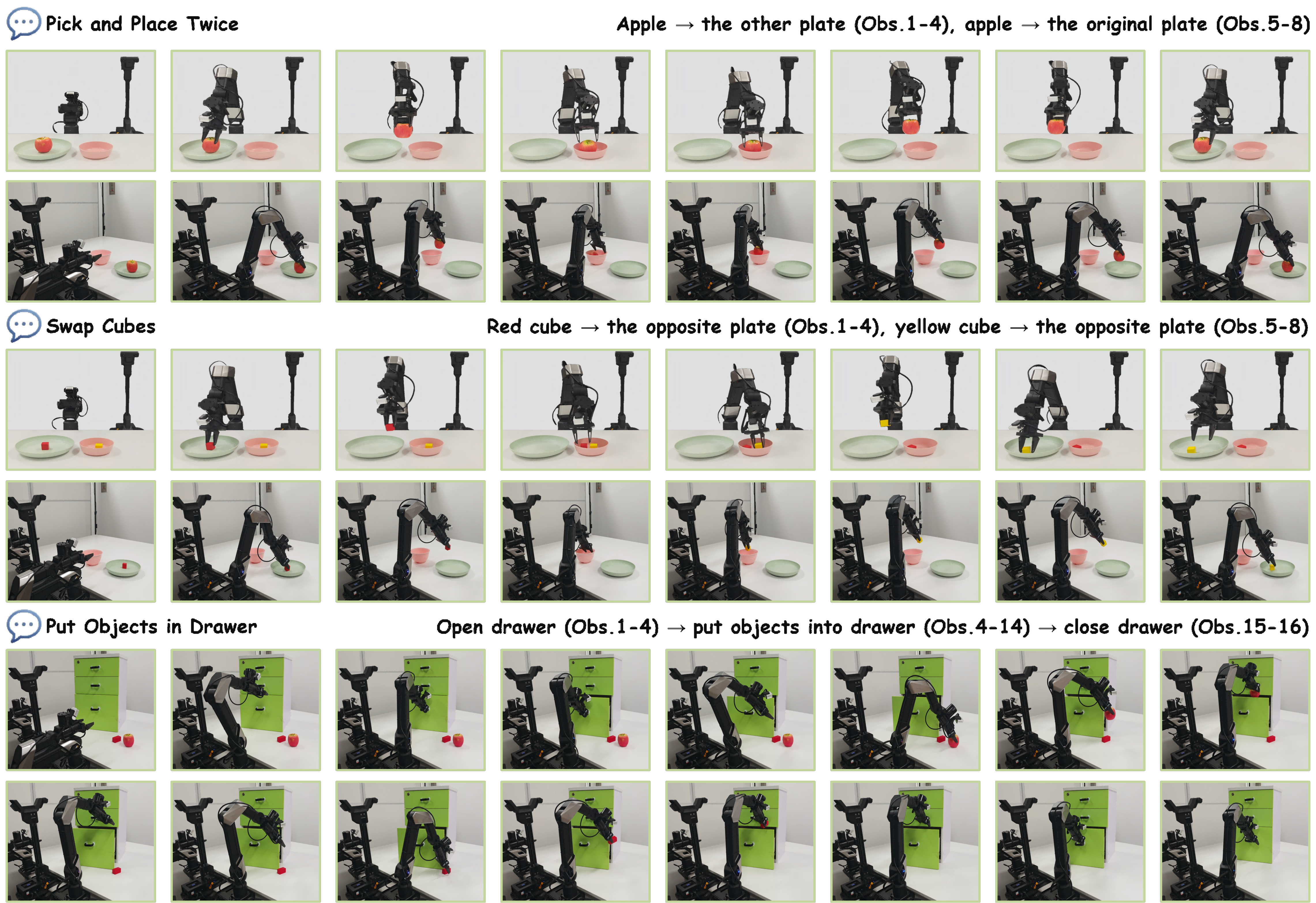}
\par\vspace{0.25em}
{\small (b) Long-horizon tasks}
\caption{Representative successful real-world rollouts. Put Objects in Drawer ends with drawer closure after both objects are inserted. The long-horizon examples highlight the need for temporally consistent stage tracking across multi-step execution.}
\label{fig:appendix_real_success_process}
\end{figure}

Representative successful real-world rollouts are shown in \cref{fig:appendix_real_success_process}. The upper panel highlights the two short-horizon tasks, Put Apple on Plate and Open then Close Drawer, while the lower panel shows Pick and Place Twice, Swap Cubes, and Put Objects in Drawer. These examples complement the real-world quantitative results in the main paper by illustrating the stage-wise behavior of the policy on representative executions.

\begin{figure}[htbp]
\centering
\includegraphics[width=0.98\linewidth]{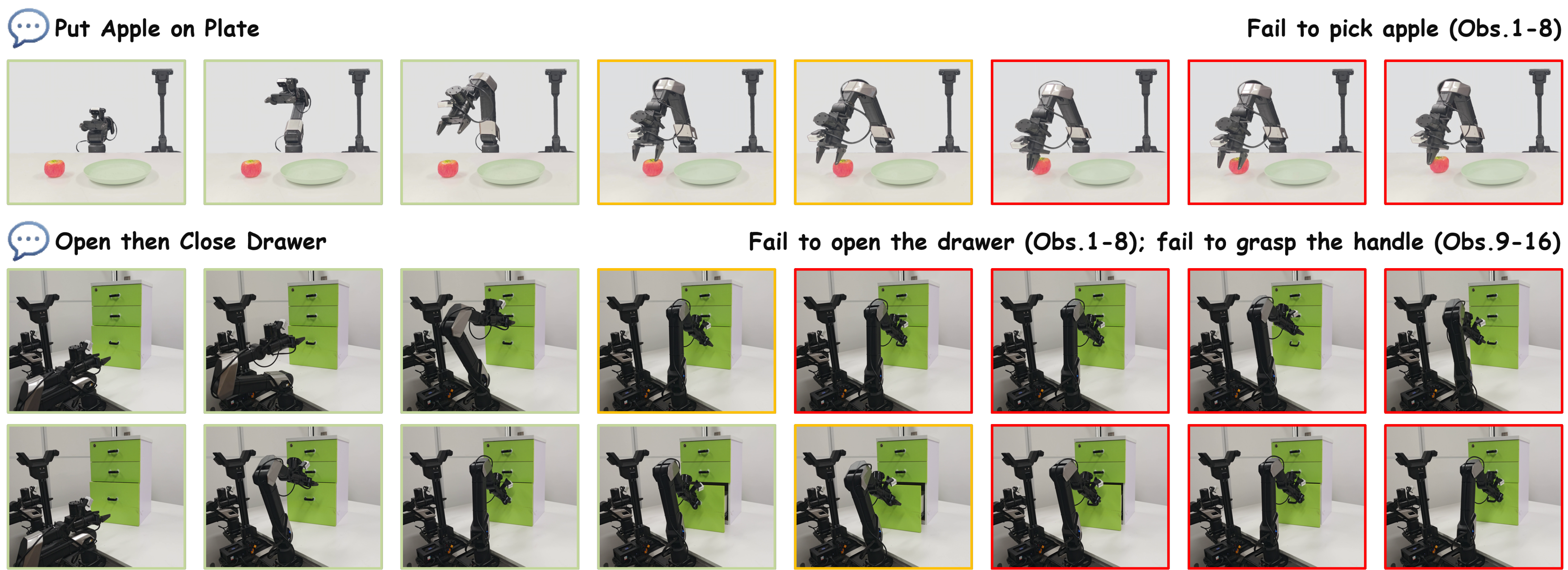}
\par\vspace{0.25em}
{\small (a) Simple-task failures.}
\par\vspace{0.8em}
\includegraphics[width=0.98\linewidth]{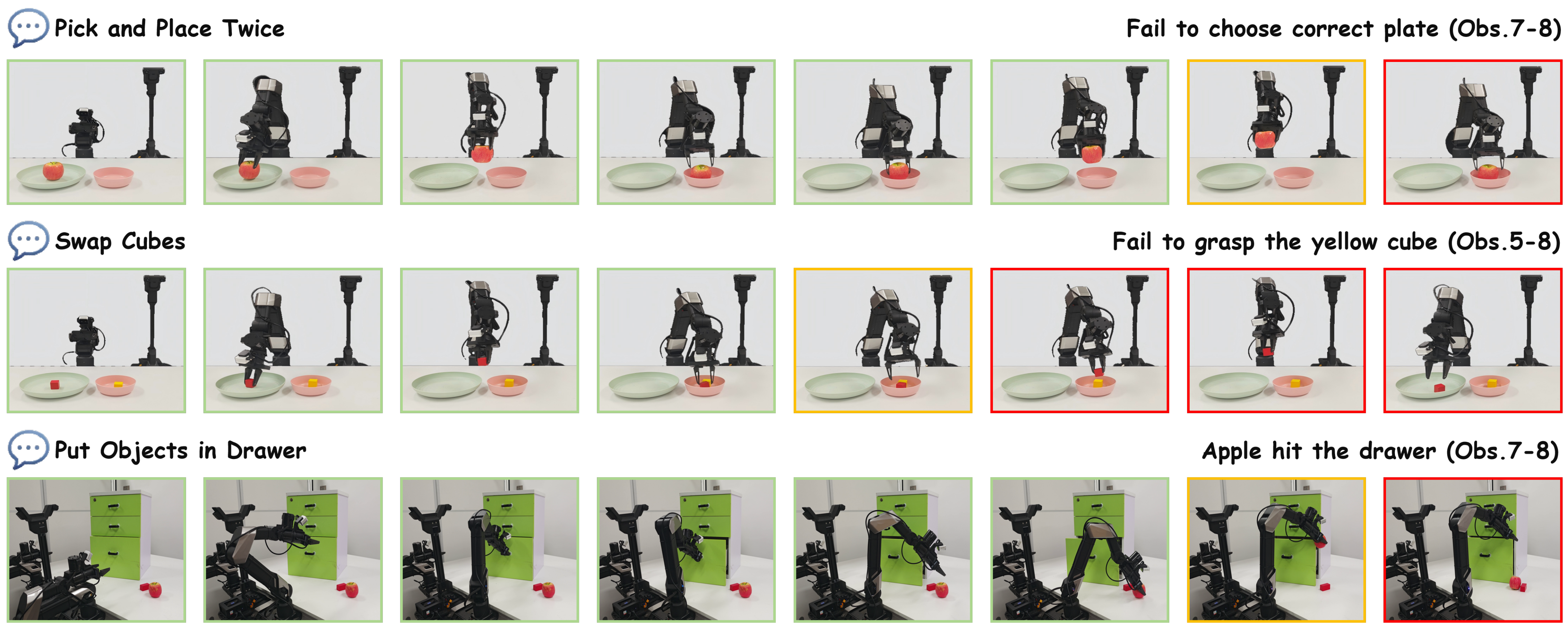}
\par\vspace{0.25em}
{\small (b) Long-horizon failures.}
\caption{Representative real-world failure cases. In the simple-task portion, Put Apple on Plate typically fails when the apple is not grasped successfully, and Open then Close Drawer fails when the policy misses the drawer handle during opening or closing, sometimes leading to direct pushing of the drawer front instead of handle-based interaction. In the long-horizon portion, Pick and Place Twice may fail by selecting the wrong target plate; Swap Cubes may fail by grasping the wrong cube and returning it; Put Objects in Drawer may fail through object--drawer collision during placement or through the same handle-grasp failure modes observed in the drawer task.}
\label{fig:appendix_real_failure_cases}
\end{figure}

Representative failure cases for the simple and long-horizon tasks are shown in \cref{fig:appendix_real_failure_cases}. In Put Apple on Plate, failure is typically caused by missing the apple during the grasp attempt. In Open then Close Drawer, failure occurs when the policy fails to grasp the handle while opening the drawer, or again fails to grasp the handle while closing it and instead pushes the drawer front directly. In Pick and Place Twice, the typical error is placing the object on the wrong plate. In Swap Cubes, the policy may grasp the wrong cube and then put it back. In Put Objects in Drawer, failure can arise either because the object collides with the drawer during placement and drops, or because the policy exhibits the same handle-grasp failure mode as Open then Close Drawer when interacting with the drawer itself.

\section{Limitations, Broader Impact, and Asset Usage}

\subsection{Limitations}
\label{sec:appendix_limitations}
\paragraph{Finite history and representation.}
\MethodName\ conditions low-level actions on a bounded recent visual window. The FIFO provides no explicit retrieval path once an event leaves the window, and the model has neither persistent semantic task state nor a high-level planner. A long execution horizon does not imply equally long explicit memory. RMBench exposes this boundary: $M(n)$ success is only 12.5\%, and Press Button has zero success. LHC also depends on the per-frame VLM representation; it cannot reconstruct task-relevant evidence that the per-frame features fail to encode. Slot-attention patterns are descriptive and do not establish fixed slot semantics or causal mechanisms.

\paragraph{Hyperparameter selection and transfer.}
History length $M$ and slot budget $K$ are fixed within each benchmark setting and interact: the Simpler grid favors $K=6$ at $M=20$, but $K=10$ at $M=30$. We selected $M=20,K=6$ using Simpler performance and transferred it unchanged to RMBench. Real-world models also use $M=20,K=6$, while LIBERO uses empirically selected $M=10,K=3$. No independent held-out hyperparameter split was used, so results on the selection tasks can be optimistic estimates of transfer performance. The RMBench experiment supports transfer of one selected configuration; it does not establish a generally robust default or an optimal compression ratio. Adaptive history budgets and independent validation remain useful future directions.

\paragraph{Statistical and experimental scope.}
Simpler uncertainty estimates and LIBERO standard deviations concern evaluation seeds of fixed trained checkpoints; real-world intervals concern the evaluated policies and deployment setting. These summaries do not quantify variability across independent training runs. Pooled Wilson intervals also do not fully represent repeated-scene dependence or task heterogeneity. Evaluation covers one OpenVLA/CronusVLA backbone family and one physical robot platform, Galaxea R1 Lite, with five physical tasks. Generalization to other backbones, embodiments, and task families remains unestablished. LHC adds substantial capacity relative to the decoder (Appendix~\ref{sec:appendix_protocol_details}), so the present evidence cannot identify temporal selection as the sole cause of improvement.

\subsection{Broader Impact}
Decoder-side condensation of bounded recent history may improve manipulation in structured environments, such as laboratory automation, where transient observations help resolve the current subtask. The measured latency scaling may also benefit research using longer conditioning windows. These benefits are limited to the evaluated policy family and do not establish general-purpose autonomy, persistent memory, or safety for deployment.

Failures can have physical consequences. Forgetting an earlier event after it leaves the FIFO, selecting the wrong subgoal, or misinterpreting contact and occlusion can lead to dropped objects, collisions, and damage to equipment or injury to nearby people. Such failures may disproportionately limit use in environments unlike the training and evaluation settings. Improved benchmark success should therefore not be treated as evidence that unsupervised operation is appropriate. Potential mitigations include task- and site-specific validation, human supervision, constrained workspaces, and independent stop mechanisms. These are deployment recommendations, not safeguards evaluated or demonstrated by this study.

Pretrained VLA components also raise model supply-chain concerns: INFUSE demonstrates that implanted backdoors can persist through downstream fine-tuning~\cite{zhou2026inject}.

\subsection{Assets and Licenses}
The paper builds on existing research assets, including OpenVLA/CronusVLA-style pretrained policy components, SimplerEnv, LIBERO, and previously reported baseline results. We cite the original papers for these assets and use them only for research and evaluation purposes under their intended terms of use. This submission does not redistribute third-party datasets, checkpoints, simulator assets, or pretrained model weights. Any future public release of code or checkpoints will include appropriate attribution and will follow the licenses and terms associated with the underlying datasets, models, and benchmark environments.

\newpage
\section*{NeurIPS Paper Checklist}

\begin{enumerate}

\item {\bf Claims}
    \item[] Question: Do the main claims made in the abstract and introduction accurately reflect the paper's contributions and scope?
    \item[] Answer: \answerYes{}.
    \item[] Justification: The abstract and introduction state the proposed learnable history-condensation module, the scoped benchmarks, and the main empirical findings. The claims are restricted to the evaluated Simpler-WidowX, LIBERO, RMBench, and real-world settings; the history mechanism is bounded by a finite FIFO.

\item {\bf Limitations}
    \item[] Question: Does the paper discuss the limitations of the work performed by the authors?
    \item[] Answer: \answerYes{}.
    \item[] Justification: The main text and appendix discuss finite FIFO history, sensitivity to history/slot budgets, selection without a held-out tuning split, fixed-checkpoint statistics, frozen visual representations, the single backbone family and physical robot platform, and the contribution of added trainable capacity.

\item {\bf Theory assumptions and proofs}
    \item[] Question: For each theoretical result, does the paper provide the full set of assumptions and a complete (and correct) proof?
    \item[] Answer: \answerNA{}.
    \item[] Justification: The paper is empirical and does not introduce formal theoretical results or proofs.

    \item {\bf Experimental result reproducibility}
    \item[] Question: Does the paper fully disclose all the information needed to reproduce the main experimental results of the paper to the extent that it affects the main claims and/or conclusions of the paper (regardless of whether the code and data are provided or not)?
    \item[] Answer: \answerYes{}.
    \item[] Justification: The main paper and appendix describe the architecture, objective, trainable modules, benchmark settings, hyperparameters, parameter counts, and evaluation protocols needed to reproduce the main experimental claims.

\item {\bf Open access to data and code}
    \item[] Question: Does the paper provide open access to the data and code, with sufficient instructions to faithfully reproduce the main experimental results, as described in supplemental material?
    \item[] Answer: \answerNo{}.
    \item[] Justification: Code and checkpoints are not released at submission time because they are not yet organized for anonymous public release. We plan to release cleaned code and reproduction instructions in a future public version.

\item {\bf Experimental setting/details}
    \item[] Question: Does the paper specify all the training and test details (e.g., data splits, hyperparameters, how they were chosen, type of optimizer) necessary to understand the results?
    \item[] Answer: \answerYes{}.
    \item[] Justification: The main paper and appendix specify the training scope, benchmark-specific data, input modalities, history and slot settings, learning rate, batch size, epochs, diffusion steps, and evaluation rollouts.

\item {\bf Experiment statistical significance}
    \item[] Question: Does the paper report error bars suitably and correctly defined or other appropriate information about the statistical significance of the experiments?
    \item[] Answer: \answerYes{}.
    \item[] Justification: The appendix defines evaluation-seed SD and reports pooled Wilson 95\% intervals for the repeated fixed-checkpoint Simpler comparison and the 50-trial real-world comparison, with the formula and binomial-model limitations. The LIBERO appendix reports CondenseVLA mean and SD across three evaluation seeds of fixed trained checkpoints; the main table reports means only. These summaries do not cover training randomness, every ablation, or task-wise statistical significance.

\item {\bf Experiments compute resources}
    \item[] Question: For each experiment, does the paper provide sufficient information on the computer resources (type of compute workers, memory, time of execution) needed to reproduce the experiments?
    \item[] Answer: \answerNo{}.
    \item[] Justification: The appendix reports A100-SXM4-80GB inference profiling with batch size 1, 10-step DDIM, warm-up, episode counts, and component timings. Per-run training hardware/time and a total research compute budget are not fully reported.
    
\item {\bf Code of ethics}
    \item[] Question: Does the research conducted in the paper conform, in every respect, with the NeurIPS Code of Ethics \url{https://neurips.cc/public/EthicsGuidelines}?
    \item[] Answer: \answerYes{}.
    \item[] Justification: The research uses standard robot-manipulation benchmarks and controlled laboratory robot experiments, and we are not aware of deviations from the NeurIPS Code of Ethics.

\item {\bf Broader impacts}
    \item[] Question: Does the paper discuss both potential positive societal impacts and negative societal impacts of the work performed?
    \item[] Answer: \answerYes{}.
    \item[] Justification: The appendix discusses possible benefits of bounded recent-history conditioning, physical harms from subgoal/contact errors and finite memory, deployment limits, and possible mitigations. Supervision, constrained workspaces, and stop mechanisms are recommendations, not safeguards validated in this study.
    
\item {\bf Safeguards}
    \item[] Question: Does the paper describe safeguards that have been put in place for responsible release of data or models that have a high risk for misuse (e.g., pre-trained language models, image generators, or scraped datasets)?
    \item[] Answer: \answerNA{}.
    \item[] Justification: The paper does not release high-risk generative models, scraped datasets, or broadly deployable autonomous robot systems as part of the submission.

\item {\bf Licenses for existing assets}
    \item[] Question: Are the creators or original owners of assets (e.g., code, data, models), used in the paper, properly credited and are the license and terms of use explicitly mentioned and properly respected?
    \item[] Answer: \answerYes{}.
    \item[] Justification: We cite the original papers for the existing models, datasets, simulators, and benchmark results used in the work. The appendix further states that third-party assets are used for research and evaluation under their intended terms and are not redistributed in this submission.

\item {\bf New assets}
    \item[] Question: Are new assets introduced in the paper well documented and is the documentation provided alongside the assets?
    \item[] Answer: \answerNA{}.
    \item[] Justification: The paper does not introduce or release a new dataset, benchmark, code package, or model checkpoint as part of this submission.

\item {\bf Crowdsourcing and research with human subjects}
    \item[] Question: For crowdsourcing experiments and research with human subjects, does the paper include the full text of instructions given to participants and screenshots, if applicable, as well as details about compensation (if any)? 
    \item[] Answer: \answerNA{}.
    \item[] Justification: The paper does not involve crowdsourcing experiments or human-subject studies.

\item {\bf Institutional review board (IRB) approvals or equivalent for research with human subjects}
    \item[] Question: Does the paper describe potential risks incurred by study participants, whether such risks were disclosed to the subjects, and whether Institutional Review Board (IRB) approvals (or an equivalent approval/review based on the requirements of your country or institution) were obtained?
    \item[] Answer: \answerNA{}.
    \item[] Justification: The paper does not involve crowdsourcing experiments or human-subject studies, so IRB approval or equivalent review is not applicable.

\item {\bf Declaration of LLM usage}
    \item[] Question: Does the paper describe the usage of LLMs if it is an important, original, or non-standard component of the core methods in this research? Note that if the LLM is used only for writing, editing, or formatting purposes and does \emph{not} impact the core methodology, scientific rigor, or originality of the research, declaration is not required.
    %this research? 
    \item[] Answer: \answerNA{}.
    \item[] Justification: The core method does not use LLMs as an important, original, or non-standard component; any LLM use limited to writing, editing, or formatting would not affect the methodology or experimental results.

\end{enumerate}

\end{document}